\documentclass{article}

\usepackage[preprint]{neurips_2021}

\usepackage[utf8]{inputenc} 
\usepackage[T1]{fontenc}    
\usepackage{hyperref}       
\usepackage{url}            
\usepackage{booktabs}       
\usepackage{amsfonts}       
\usepackage{amsmath}
\usepackage{mathtools}
\usepackage{nicefrac}       
\usepackage{microtype}      
\usepackage{xcolor}         
\usepackage{multirow}
\usepackage{graphicx}
\usepackage{bm}
\usepackage{rotating}

\newcommand{\ra}[1]{\renewcommand{\arraystretch}{#1}}

\DeclareMathOperator*{\argmin}{arg\,min}

\newcommand{\norm}[1]{\left\lVert#1\right\rVert}

\title{Convolutional Motif Kernel Networks}

\author{%
  \hspace{1mm}Jonas C.~Ditz \\
	Methods in Medical Informatics\\
	Department of Computer Science\\
	University of T\"{u}bingen\\
	T\"{u}bingen, Germany \\
	\texttt{jonas.ditz@uni-tuebingen.de} \\
	\And
	\hspace{1mm}Bernhard Reuter \\
	Methods in Medical Informatics\\
	Department of Computer Science\\
	University of T\"{u}bingen\\
	T\"{u}bingen, Germany \\
	\texttt{bernhard.reuter@uni-tuebingen.de} \\
	\And
	\hspace{1mm}Nico Pfeifer \\
	Methods in Medical Informatics\\
	Department of Computer Science\\
	University of T\"{u}bingen\\
	T\"{u}bingen, Germany \\
	\texttt{nico.pfeifer@uni-tuebingen.de} \\
}

\begin{document}

\maketitle

\begin{abstract}
  Artificial neural networks show promising performance in detecting correlations within data that are associated with specific outcomes. However, the black-box nature of such models can hinder the knowledge advancement in research fields by obscuring the decision process and preventing scientist to fully conceptualize predicted outcomes. Furthermore, domain experts like healthcare providers need explainable predictions to assess whether a predicted outcome can be trusted in high stakes scenarios and to help them integrating a model into their own routine. Therefore, interpretable models play a crucial role for the incorporation of machine learning into high stakes scenarios like healthcare. In this paper we introduce Convolutional Motif Kernel Networks, a neural network architecture that involves learning a feature representation within a subspace of the reproducing kernel Hilbert space of the position-aware motif kernel function. The resulting model enables to directly interpret and evaluate prediction outcomes by providing a biologically and medically meaningful explanation without the need for additional \textit{post-hoc} analysis. We show that our model is able to robustly learn on small datasets and reaches state-of-the-art performance on relevant healthcare prediction tasks. Our proposed method can be utilized on DNA and protein sequences. Furthermore, we show that the proposed method learns biologically meaningful concepts directly from data using an end-to-end learning scheme.
\end{abstract}

\section{Introduction}
Biological sequences contain valuable information for a wide variety of biological processes. While this property makes them crucial for advances in related research fields, it also provides the potential to improve diagnosis and treatment decisions in healthcare systems. For this reason, a large amount of machine learning approaches that solve learning tasks on biological sequences were developed over the last years. Among others, these approaches include the prediction of splice sites \citep{degroeve2002feature} and translation initiation sites \citep{zien2000engineering}, predicting binding affinity between proteins and DNA/RNA \citep{alipanahi2015predicting,yang2020predba}, drug resistance prediction \citep{doring2018geno2pheno}, or the denoising of biological sequence data \citep{eraslan2019single}. However, trained models can only be safely incorporated into medical routines if their prediction outcomes can be thoroughly interpreted and understood even by domain experts, e.g., healthcare providers like medical practitioners, without strong knowledge in the foundations of machine learning. Kernel methods and statistical models provide the possibility to interpret results within the data's domain, hence, allowing domain experts to judge outcomes using their own expertise. Yet, scalability issues in terms of data size limit their utility considering the rapid increase of available data in medical and biological research. On the other hand, gradient-based learning approaches like neural networks can handle huge data pools with relative ease but are normally developed as black-box models. Although there are model-agnostic techniques to interpret these models, e.g., saliency maps \citep{kadir2001saliency} or Shapley additive explanations (SHAP) \citep{lundberg2017unified}, recent work by Rudin \citep{rudin2019stop} advises the use of inherently interpretable models for high stakes scenarios over \textit{post-hoc} explaining black-box models. One problem of \textit{post-hoc} ML explanation models identified by Rudin is their unfaithfulness regarding the original model's computation, which can result in misleading explanations. Sixt and colleagues showed this unfaithfulness for attribution methods by proving that most methods ignored later layers of a model when computing explanations \citep{sixt2020explanations}. Furthermore, Bordt and colleagues showed the limitations of \textit{post-hoc} explanations in adversarial contexts \citep{bordt2022post}. Lipton warned about the danger of optimizing \textit{post-hoc} methods to produce plausible but misleading explanations \citep{lipton2018mythos}. In high stakes scenarios like healthcare, decisions made on misleading or wrong explanations can cause dangerous situations with the potential to further harm patients or other vulnerable groups.

In recent years, several efforts were published to combine kernel functions and neural networks \citep{cho2009kernel,bo2011object,mairal2014convolutional,mairal2016end}. Combining these two approaches enhances neural networks with the interpretability and robustness of kernel methods. On the other hand, it allows to extend learning within a reproducing kernel Hilbert space (RKHS) to problems with massive numbers of data points. Recently, Chen and colleagues introduced these efforts into data mining on biological sequences by developing convolutional kernel networks based on a continuous relaxation of the mismatch kernel \citep{chen2019biological}. Although these models show promising performance, the choice of kernel resulted in the necessity of a \textit{post-hoc} model for interpretation. Another limitation results from the fact that the mismatch kernel restricts considered $k$-mer occurrences to a position-independent representation \citep{eskin2003mismatch,leslie2004fast}. In many medical tasks, however, positional and compositional variability provide key information. One kernel network approach that utilizes positional information is the recurrent kernel network (RKN) proposed by Chen and colleagues\citep{chen2019recurrent}. While this architecture showed promising performance capabilities, the chosen architecture resulted once again in a black-box models with the need for \textit{post-hoc} interpretation.
The oligo kernel proposed by Meinicke and colleagues is able to model positional variability and can additionally provide traditional monomer-based representations as well as position-independent $k$-mer representations as limiting cases \citep{meinicke2004oligo}. Furthermore, the oligo kernel allows for intuitive and simple interpretation of $k$-mer relevance and positional variability. However, the oligo kernel cannot be directly incorporated into a convolutional network architecture and does not take into account information provided by compositional variability of motifs. While $k$-mers are short sequences with fixed letters at each position, motifs are short sequence patterns that can represent more than one possible letter at each position. The above mentioned limitations motivated our work presented here.

This work is structured in the following way. Section \ref{sec:methods} introduces the position-aware motif kernel function and details how to incorporate the position-aware motif kernel into a convolutional kernel layer and how to interpret a trained CMKN model. Section \ref{sec:experiments} provides details regarding the conducted experiments on synthetic and real-world data and the results. Finally, section \ref{sec:discussion} provides a discussion of presented prediction and interpretation results and section \ref{sec:conclusion} completes this work with a conclusion.

In summary, our manuscript provides the following contributions:
\begin{itemize}
	\item We extend convolutional kernel network models for biological sequences to incorporate positional information and make them inherently interpretable, which removes the necessity for \textit{post-hoc} explanation models. The new models are called convolutional motif kernel networks (CMKNs).
	\item This extension is achieved by introducing a new kernel function, called position-aware motif kernel, that quantifies the position dependent similarity of motif occurrences.
	\item We use one synthetic and two real-world datasets to show how our method can be used as a research tool to gain insight into biological sequence data and how CMKNs can provide local interpretation that can help domain experts, e.g., healthcare providers, to quickly interpret and validate prediction outcomes of a trained CMKN model with their domain expertise.
\end{itemize}


\section{Methods}\label{sec:methods}
In the following section, we will introduce our new kernel function and show how this kernel can be used to create inherently interpretable kernel networks.

\subsection{Position-Aware Motif Kernel}\label{sec:motif_kernel}
We introduce a new kernel function that incorporates the positional uncertainty of the oligo kernel \citep{meinicke2004oligo} but is defined for arbitrary sequence motifs. Furthermore, our kernel function can be used to construct a convolutional kernel layer as described by Mairal \citep{mairal2016end}. Our kernel function is based on two main ideas: First, we introduce a mapping of sequence positions onto the unit circle, which allows us to represent the position comparison term by a linear operation followed by a non-linear activation function. Second, we introduce a $k$-mer comparison term. This extension enables the kernel function to deal with inexact $k$-mer matching, which capacitates our kernel function to handle arbitrary sequence motifs. We call our new kernel function position-aware motif kernel (PAM).

The first part of our position-aware motif kernel compares sequence positions. In prior work, e.g., Meinicke et~al., 2004 \citep{meinicke2004oligo} or Mialon et~al., 2021 \citep{mialon2021atrainable}, a quadratic term is usually employed to measure the similarity of positions. We utilize a linear comparison term instead. First, all positions are mapped onto the upper half of the unit circle to create unit $\ell_2$-norm vectors: $\tilde{p} = \left( (\textrm{cos}\left( \frac{p}{|\mathbf{x}|} \pi \right), \textrm{sin}\left( \frac{p}{|\mathbf{x}|} \pi \right) \right)^T,$ where $|\mathbf{x}|$ denotes the length of the corresponding sequence. Due to the position vectors now having unit $\ell_2$-norm, the position comparison term can be written as follows: $-\frac{1}{4\sigma} \norm{\tilde{p} - \tilde{q}}_2^2 = \frac{1}{2\sigma} (\tilde{p}^T \tilde{q} - 1)$. This allows us to define the following position comparison kernel function over pairs of sequence positions:
\begin{equation}
\label{eqn:poskernel}
K_{\textrm{position}}(p, q) = \exp \left( \frac{\beta}{2 \sigma^2} \left( \tilde{p}^{T} \tilde{q} - 1 \right) \right),
\end{equation}
where $\beta$ is a scaling parameter that compensates for the reduced absolute distance between sequence positions due to the introduced mapping and $\sigma$ is a positional uncertainty parameter similar to the homonymous $\sigma$ parameter of the oligo kernel.

The second part of our position-aware motif kernel compares sequence motifs. For biological sequences, a motif describes a nucleotide or amino acid pattern of a certain length. Sequence motifs can be written in form of a normalized position frequency matrix (nPFM), which is a matrix in $\mathbb{R}_{+}^{|\mathbf{A}| \times k}$ with $|\mathbf{A}|$ being the size of the alphabet over which the motif is created and $k$ being the length of the motif. An nPFM has to fulfill the additional constraint that each column has unit $\ell_2$-norm (see supplement for more details). For two motifs $\omega$ and $\omega'$ of length $k$ given as flattened nPFMs, i.e., the columns are concatenated to convert the matrix into a vector, we define the following motif comparison kernel function:
\begin{equation}
\label{eqn:olikernel}
K_{\textrm{nPFM}}(\omega, \omega') = \exp \left( \alpha \left( \omega^{T} \omega' - k \right) \right).
\end{equation}
This function will become one if the two motifs match exactly and will approach zero with increasing difference of the two motifs. The parameter $\alpha$ determines how fast the function approaches zero and, hence, specifies the influence of inexact matching motifs.

We define our position-aware motif kernel by forming the product kernel using the functions introduced in Equation \ref{eqn:poskernel} and \ref{eqn:olikernel} and aggregating the kernel evaluation of all motif-position pairs with a sum. In other words, the position-aware motif kernel for pairs of sequences $\mathbf{x}$ and $\mathbf{x}'$ over an alphabet $\mathbf{A}$ is given by:
\begin{equation}
\label{eqn:cok}
K_{\textrm{PAM}}(\mathbf{x}, \mathbf{x}') =
C \sum_{p = 1}^{|\mathbf{x}|} \sum_{q = 1}^{|\mathbf{x'}|} K_0((\mathbf{\omega}_{p}, p), (\mathbf{\omega}_{q}, q))
\end{equation}
with
\begin{align}
\label{eqn:kzero}
K_0((\omega_{p}, p), (\omega_{q}, q)) = K_{\textrm{nPFM}}(\omega_{p}, \omega_{q}) \cdot K_{\textrm{position}}(p, q) = \exp \left( \alpha \left( \omega^{T}_{p} \omega_{q} - k \right) + \frac{\beta}{2 \sigma^2} \left( \tilde{p}^{T} \tilde{q} - 1 \right) \right). \nonumber
\end{align}
Here, $|\mathbf{x}|$ and $|\mathbf{x}'|$ are the lengths of the respective sequences, $\omega_{p}$ is the motif of length $k$ starting at position $p$ in sequence $\mathbf{x}$ represented as a flattened nPFM, and $\omega_{q}$ is defined analogously to $\omega_{p}$ but for sequence $\mathbf{x}'$. The constant $C=\sqrt{\frac{\pi^2\sigma^2}{2\alpha\beta}}$ results from the derivation of the motif kernel matrix elements as the inner product of two sequence representatives $\phi_{\mathbf{x}}, \phi_{\mathbf{x}'}$ in the feature space of all motifs as detailed in the Supplement.

\subsection{Extracting a Feasible Kernel Layer using Nystr\"{o}m's Method}\label{sec:nystroem}

\begin{figure}
    \centering
    \includegraphics[width=0.7\textwidth]{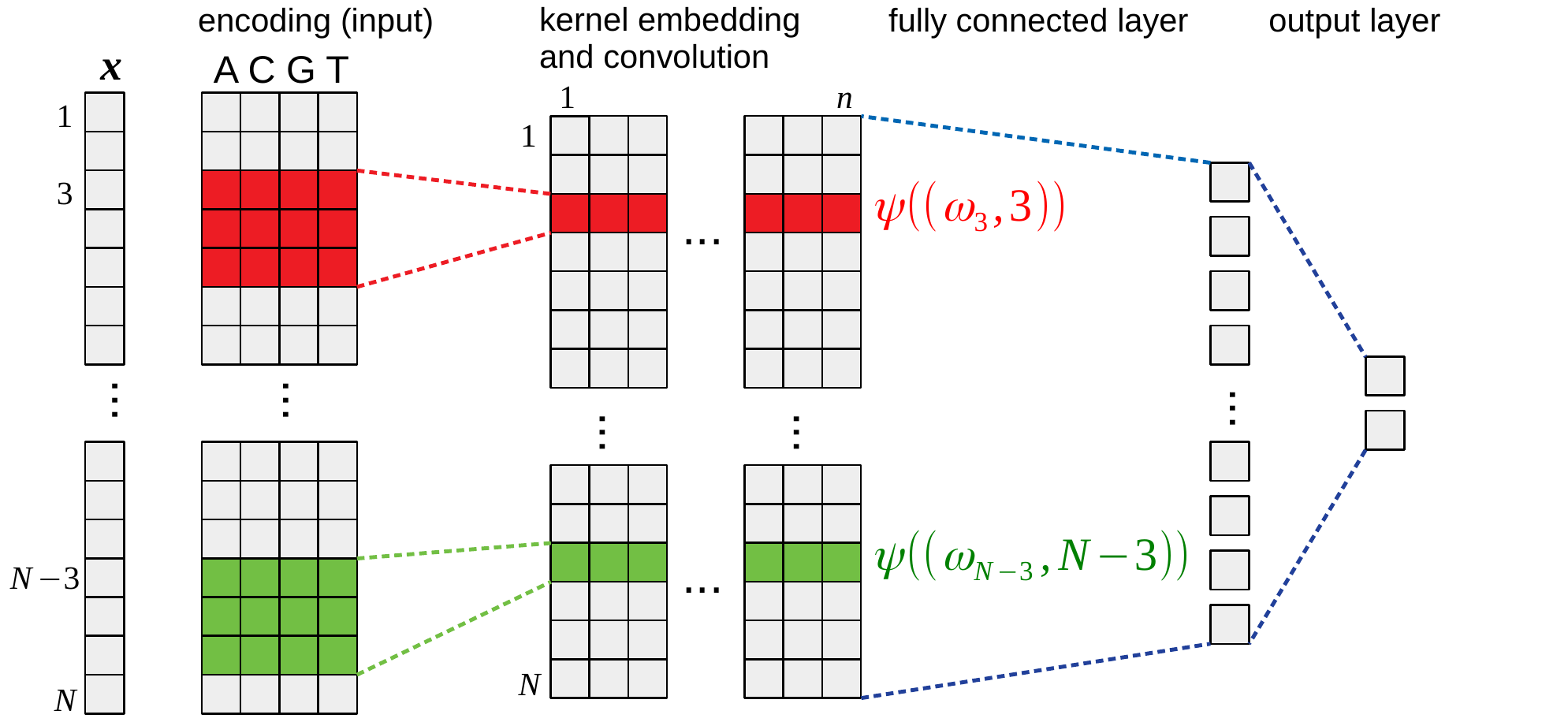}
    \caption{Schematic overview of an CMKN model. Each motif-position pair of the input is projected onto the subspace of the RKHS by the kernel layer. Afterwards, the projected input is classified using one or several linear fully-connected layers.}
    \label{fig:architecture}
\end{figure}

Mairal and colleagues showed that a variant of the Nystr\"{o}m method \citep{williams2001using,zhang2008improved} can be used to incorporate learning within a reproducing kernel Hilbert space (RKHS) into neural networks \citep{mairal2014convolutional, mairal2016end}. We use the same approach to construct a finite-dimensional subspace of the RKHS $\mathcal{H}$ over motif-position pairs that is implicitly defined by $K_0$ and incorporate learning within this subspace into a neural network architecture.

Consider a set of $n$ anchor points $z_1, \ldots, z_n$, where each anchor point is a motif-position pair $z_i = \left( \omega_{z_i}, p_{z_i} \right)$. We define an $n$-dimensional subspace $\mathcal{E}$ of $\mathcal{H}$ that is spanned by a set of anchor points, i.e.
\begin{equation}
\mathcal{E} = \textrm{Span}\left( \phi_{z_1}, \ldots, \phi_{z_n} \right),
\end{equation}
where $\phi_{z_i}$ denotes the projection of each anchor point into the RKHS $\mathcal{H}$. Utilizing the kernel trick, a motif-position pair can be projected onto $\mathcal{E}$ without explicitly calculating the images of the anchor points $\phi_{z_1}, \ldots, \phi_{z_n}$. This natural parametrization is given by \citep{mairal2016end}
\begin{equation}
\label{eqn:mapping}
\psi((\omega, p)) = K_{ZZ}^{-\frac{1}{2}} K_{Z}((\omega, p)).
\end{equation}
Here, $K_{ZZ}=(K_0(z_i, z_j))_{i=1,\ldots,n;j=1,\ldots,n}$ is the Gram matrix formed over the anchor points, $K_{ZZ}^{-\frac{1}{2}}$ is the (pseudo)-inverse square root of the Gram matrix, and $K_{Z}((\omega, p)) = (K_0(z_1, (\omega, p)), \ldots, K_0(z_n, (\omega, p)))^T$. We follow the procedure proposed in prior work \citep{mairal2016end,chen2019biological} to initialize anchor points. First, we sample a set of $m >> n$ motif-position pairs from the training data. Afterwards, we perform k-means clustering with euclidean distance metric using k-means++ initialization to get $n$ cluster centers of the sampled set. After convergence, we enforce the nPFM constraints onto the cluster centers. With initialized anchor points, CMKN models can be trained by a simple end-to-end learning scheme. A schematic overview over a CMKN model for DNA input together with a visualization of the information flow within the network is shown in Figure \ref{fig:architecture}.

\subsection{Interpreting a CMKN Model}\label{sec:viz}
The main intuition behind the position-aware motif kernel is to detect similarities between motifs, even if they occur at a certain distance from each other and even if the nPFM underlying the motif is different to a certain degree. In this way, our approach extends previous approaches like the oligo kernel \citep{meinicke2004oligo} and the weighted degree kernel with shifts \citep{Ratsch2005}, which only evaluated exact $k$-mer matches. However, our kernel is based on a concept that we call motif functions which are extensions of the oligo functions introduced by Meinicke and colleagues \citep{meinicke2004oligo} (see Supplement for details). A motif function represents the nPFM and position(s) of occurrence of the corresponding motif with a smoothing of the position to account for positional uncertainty. Apart from providing a biologically meaningful feature representation, the use of a kernel based on motif functions allows for a direct interpretation of a trained CMKN model without the need for \textit{post-hoc} methods. If the CMKN model consists only of linear fully-connected layers after the kernel layer, as strictly applied throughout this study, important sequence positions and corresponding motifs can be directly inferred from the learned weights and anchor points, since this ensures that only linear combinations of the learned feature representations are considered. The importance of a sequence position for a certain class can be assessed by calculating the mean positive weight of the edges that connect the position with the output state that corresponds to the class. The importance $\iota$ of position $p$ for class $c$ can thereby be expressed as:
\begin{equation}
    \iota^p_c = \frac{1}{|N_p|} \sum_{n \in N_p} \Tilde{\iota}_{n, c} \, \, , \qquad \Tilde{\iota}_{n, c} = \begin{cases}
        \sum_{\{m | m \in N^{(n)}\}} w_{n, m} \Tilde{\iota}_{m, c}, & \textrm{if } N^{(n)} \cap N^{(O)} = \emptyset\\
        1, & \textrm{if } N^{(n)} \cap N^{(O)} = o_c \\
        0, & \textrm{otherwise}.
    \end{cases}
\end{equation}
Here, $N_p$ denotes the set of neurons contributing to the importance of position $p$, $N^{(n)}$ denotes the set of neurons from the next layer connected by an edge with positive weight to neuron $n \in N_p$, $w_{n, m}$ denotes the weight of the edge connecting neuron $n$ with neuron $m\in N^{(n)}$, and $N^{(O)}$ denotes the set of $|c|$ neurons $o_c$, each representing a single class, in the output layer. Furthermore, the motif associated with the class at that position is retrieved by identifying all learned motifs with positive weights and calculating the weighted mean motif using the learned weights. This procedure is similar to inferring feature importances from the primal representation using the learned parameters of a SVM. Said utilization of the primal representation is possible for linear kernels and most string kernels \citep{meinicke2004oligo}. The importance of each amino acid at each position of the motif can be directly accessed by sorting the rows of each column of the associated nPFM in decreasing order. Additionally, motif functions enhance CMKN models with the ability to compute local interpretations, i.e., an explanation of prediction results for single inputs within the data's domain. For an input sequence and a learned motif-position pair, we can estimate the importance of that pair by calculating the $\ell_2$-norm of the corresponding motif function. To assess the class that a model associates with an important position, the class-specific motifs that were learned by a model at that position can be retrieved and ranked by the $\ell_2$-norm of the motif functions on the input sequence. The motif with the highest $\ell_2$-norm determines which class a model assigns to the position. We show an exemplary visualization for domain experts of this procedure in Figure \ref{fig:interpretability}\textbf{b}.

\section{Experiments}\label{sec:experiments}
We used synthetic data to evaluate CMKN's ability to recover meaningful sequence patterns. Furthermore, we evaluated the performance capability of our proposed method on two different prediction tasks: antiretroviral drug resistance prediction and splice site recognition.

\subsection{Recovering Meaningful Patterns in Synthetic Data}
\begin{figure}
    \centering
    \includegraphics[width=\textwidth]{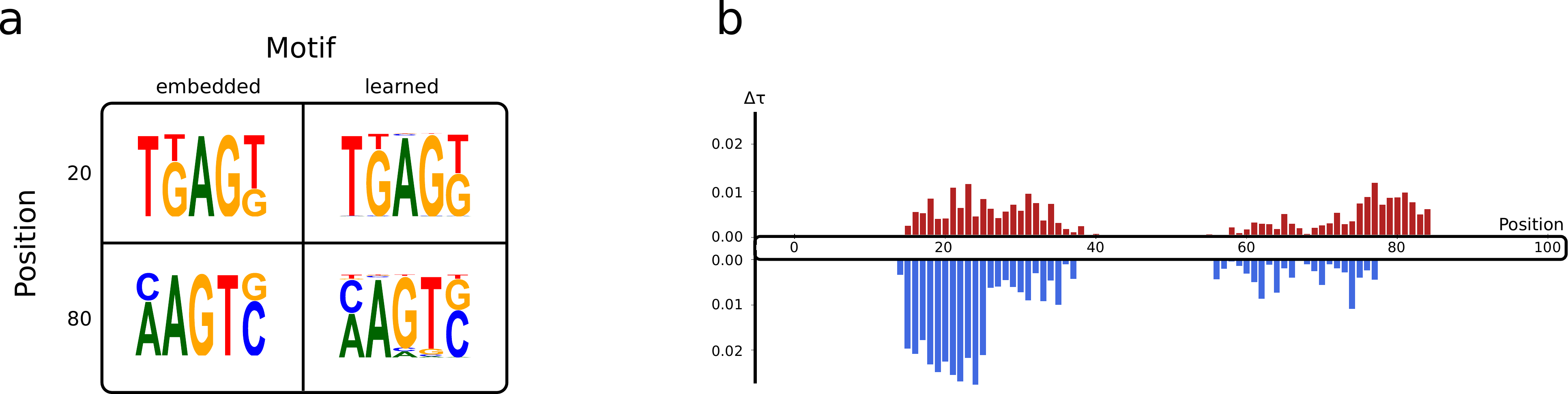}
    \caption{Evaluation of the interpretation capabilities of CMKN using synthetic data. \textbf{a}: The matrix shows the embedded motifs (left column) and the motifs learned by CMKN (right column). The first row shows the motif at position 20 which was only embedded into negative sequences. The second row shows the motif at position 80 which was only embedded into positive sequences. \textbf{b}: Positional feature importance of CMKN on the synthetic data. Each bar shows the derivation from the mean positional feature importance for the corresponding sequence position. Red bars indicate importance for the positive class and blue bars indicate importance for the negative class.}
    \label{fig:interpretation_synthetic}
\end{figure}

In order to assess whether CMKN models can reliably recover distinct biological patterns from sequences, we created a synthetic dataset containing 1000 randomly generated DNA sequences of length 100. The set was equally split into negative and positive sequences, with a distinct motif embedded into each class of sequences at a specific position (see Figure \ref{fig:interpretation_synthetic}\textbf{a} for the embedded motifs). For negative sequences, the motif was embedded at position 20 with a positional uncertainty of $\pm$ 5 positions. For positive sequences, the motif was embedded at position 80 with a positional uncertainty of $\pm$ 5 positions. The compositional variablity shown in Figure \ref{fig:interpretation_synthetic}\textbf{a} can be understood in a way that one-third of the 5-mers embedded into negative sequences had a thymine at position 2 while two-third of the k-mers had a cytosine. This is equivalent for the other motif positions with compositional variability. By creating a synthetic dataset with this procedure, we made sure that the data contains positional and compositional variability that are important for the prediction task. We trained a CMKN model using a motif length of 5 and a positional uncertainty parameter of 4. For the kernel layer, we chose 50 anchor points. The other kernel hyperparameters were set to $\alpha = 1$ and $\beta = 1000$. The model was trained for 50 epochs using the binary cross-entropy with logits loss function.

Figure \ref{fig:interpretation_synthetic} shows the results of our experiment with the synthetic dataset. We recovered the positional feature importance values as well as the learned motifs at position 20 and 80 using the procedure described in section \ref{sec:viz}. As clearly visible on the left side, CMKN is able to recover the two embedded motifs with high similarity using simple end-to-end learning without post-hoc model optimization. Furthermore, the right side of Figure \ref{fig:interpretation_synthetic} shows that CMKN is able to detect the relevant areas of biological sequences.

\subsection{Prediction of antiretroviral drug resistance}

\begin{table*}
	\centering
	\caption{Mean performance and standard derivation of prediction models for three different HIV drug classes: PIs, NRTIs, NNRTIs. Models include polynomial kernel SVMs ($\textrm{SVM}_{\textrm{poly}}$), oligo kernel SVMs ($\textrm{SVM}_{\textrm{oligo}}$), random forests (RF), convolutional neural networks (CNN), convolutional kernel networks ($\textrm{CKN}_{\textrm{seq}}$), and convolutional motif kernel networks (CMKN). Highest values are displayed in bold.}
	\small
	\begin{tabular}{crcrllll}
		\toprule
		\textbf{Drug Class} & \phantom{a} & \textbf{Model} & \phantom{a} & \textbf{Accuracy} & \textbf{F1 Score} & \textbf{auROC} & \textbf{MCC} \\
		\midrule
		PI \\
		&& $\textrm{SVM}_{\textrm{poly}}$ && 0.90 $\pm$ 0.04 & 0.83 $\pm$ 0.09 & 0.95 $\pm$ 0.03 & 0.75 $\pm$ 0.10 \\
		&& $\textrm{SVM}_{\textrm{oligo}}$ && \textbf{0.92 $\pm$ 0.03} & 0.86 $\pm$ 0.09 & \textbf{0.97 $\pm$ 0.03} & \textbf{0.81 $\pm$ 0.09} \\
		&& RF && \textbf{0.92 $\pm$ 0.04} & 0.85 $\pm$ 0.13 & \textbf{0.97 $\pm$ 0.03} & 0.79 $\pm$ 0.13 \\
		&& CNN && 0.91 $\pm$ 0.3 & 0.84 $\pm$ 0.11 & 0.94 $\pm$ 0.05 & 0.77 $\pm$ 0.11 \\
		&& $\textrm{CKN}_{\textrm{seq}}$ && 0.84 $\pm$ 0.05 &  0.72 $\pm$ 0.12 & 0.88 $\pm$ 0.05 & 0.60 $\pm$ 0.11 \\
		&& CMKN && \textbf{0.92 $\pm$ 0.03} & \textbf{0.87 $\pm$ 0.09} & 0.96 $\pm$ 0.03 & \textbf{0.81 $\pm$ 0.10} \\
		NRTI \\
		&& $\textrm{SVM}_{\textrm{poly}}$ && 0.86 $\pm$ 0.06 & 0.82 $\pm$ 0.09 & 0.90 $\pm$ 0.05 & 0.70 $\pm$ 0.12 \\
		&& $\textrm{SVM}_{\textrm{oligo}}$ && 0.88 $\pm$ 0.05 & 0.85 $\pm$ 0.09 & \textbf{0.94 $\pm$ 0.03} & 0.75 $\pm$ 0.10 \\
		&& RF && 0.88 $\pm$ 0.06 & 0.84 $\pm$ 0.12 & \textbf{0.94 $\pm$ 0.04} & 0.74 $\pm$ 0.15 \\
		&& CNN && 0.88 $\pm$ 0.05 & 0.85 $\pm$ 0.09 & 0.93 $\pm$ 0.04 & 0.74 $\pm$ 0.12 \\
		&& $\textrm{CKN}_{\textrm{seq}}$ && 0.79 $\pm$ 0.06 & 0.73 $\pm$ 0.12 & 0.85 $\pm$ 0.05 & 0.54 $\pm$ 0.13 \\
		&& CMKN && \textbf{0.89 $\pm$ 0.05} & \textbf{0.86 $\pm$ 0.09} & 0.93 $\pm$ 0.05 & \textbf{0.76 $\pm$ 0.11} \\
		NNRTI \\
		&& $\textrm{SVM}_{\textrm{poly}}$ && 0.82 $\pm$ 0.06 & 0.76 $\pm$ 0.11 & 0.84 $\pm$ 0.06 & 0.63 $\pm$ 0.14 \\
		&& $\textrm{SVM}_{\textrm{oligo}}$ && 0.89 $\pm$ 0.05 & 0.86 $\pm$ 0.11 & 0.94 $\pm$ 0.05 & 0.79 $\pm$ 0.12 \\
		&& RF && 0.88 $\pm$ 0.05 & 0.85 $\pm$ 0.09 & 0.93 $\pm$ 0.07 & 0.75 $\pm$ 0.12 \\
		&& CNN && 0.89 $\pm$ 0.04 & 0.86 $\pm$ 0.08 & 0.94 $\pm$ 0.06 & 0.78 $\pm$ 0.10 \\
		&& $\textrm{CKN}_{\textrm{seq}}$ && 0.73 $\pm$ 0.06 & 0.63 $\pm$ 0.16 & 0.78 $\pm$ 0.08 & 0.42 $\pm$ 0.15 \\
		&& CMKN && \textbf{0.91 $\pm$ 0.03} & \textbf{0.89 $\pm$ 0.06} & \textbf{0.95 $\pm$ 0.05} & \textbf{0.81 $\pm$ 0.08} \\
		\bottomrule
	\end{tabular}
	\label{tab:mean_performance}
\end{table*}

When choosing a personalized treatment combination for HIV-infected people, it is crucial to know the resistance profile of the viral variants against available drugs. It has been shown that the genetic sequence of a virus can be used to predict resistance against certain antiretroviral drugs \citep{doring2018geno2pheno}. We performed resistance prediction for drugs representing the three most commonly used antiretroviral drug classes against HIV infections: Nucleoside reverse-transcriptase inhibitors (NRTIs), non-nucleoside reverse-transcriptase inhibitors (NNRTIs), and protease inhibitors (PIs). This prediction task was chosen for evaluation of the proposed method, since it remains an highly important problem in the treatment of HIV infections and the acquired immune deficiency syndrome (AIDS) and is often considered as a role model for precision medicine.

Amino acid sequences of virus protein variants with corresponding drug resistance information were extracted from Stanford University's HIV drug resistance database (HIVdb) \citep{rhee2003human,shafer2006rationale}. An overview of the available data for each of the drugs included in the evaluation can be found in the Supplement. The network architecture used for HIV drug resistance prediction consists of a single convolutional motif kernel layer followed by two fully-connected layers. The first fully-connected layer projected the flattened output of the kernel layer onto 200 nodes and the second fully-connected layer had two output states, one for the susceptible class and one for the resistant class. The motif length and the hyperparameter $\alpha$ of the kernel function were both fixed to 1 based on prior biological knowledge (for details see supplement material). The scaling hyperparameter $\beta$ was fixed to $\tfrac{|\mathbf{x}|^2}{10}$ with $|\mathbf{x}| = 99$ for PI datasets and $|\mathbf{x}| = 240$ for NRTI/NNRTI datasets. This compensates for the transformation of sequence positions (for details see supplement material). The number of anchor points and the positional uncertainty parameter $\sigma$ were optimized using a grid search (for details see supplement material). Due to the limited number of available samples, each model was trained using a 5-fold stratified cross-validation. The data splits for each fold were fixed across models to ensure the same training environment for each hyperparameter combination. Training success was evaluated using the performance measures accuracy, F1 score, and area under the receiver operating characteristic curve (auROC). Due to the fact that some datasets were highly unbalanced, we also included the Matthew's correlation coefficient (MCC) \citep{matthews1975comparison} in the performance assessment.

Mean performances achieved for each of the three investigated drug classes can be found in Table \ref{tab:mean_performance}. Our method was able to achieve high accuracy, F1 score, and auROC values for each drug class. Even though the classification problem is highly imbalanced for some of the tested drugs, our model is still able to achieve a high Matthew's correlation coefficient (MCC) value with mean MCC performance exceeding 0.75 for each of the three investigated drug classes. We compared CMKN's performance to previously used models for HIV drug resistance prediction: SVMs with polynomial kernel \citep{doring2018geno2pheno} and random forest (RF) classifiers \citep{raposo2020random}. Furthermore, we included a SVM utilizing the oligo kernel and the $\textrm{CKN}_{\textrm{seq}}$ model \citep{chen2019biological} into our analysis. Additionally, we performed an ablation test by replacing the kernel layer with a standard convolutional layer to investigate the influence of our kernel architecture onto prediction performance (denoted by CNN in Table \ref{tab:mean_performance}). The results for all models can be found in Table \ref{tab:mean_performance}. Our method either outperformed the competitors or achieved similar performance.

\subsubsection{Utilizing CMKN's interpretation capabilities to identify resistance mutation positions and motifs}
\label{sec:res_viz}

\begin{figure*}[t!]
    \centering
    \includegraphics[width=\textwidth]{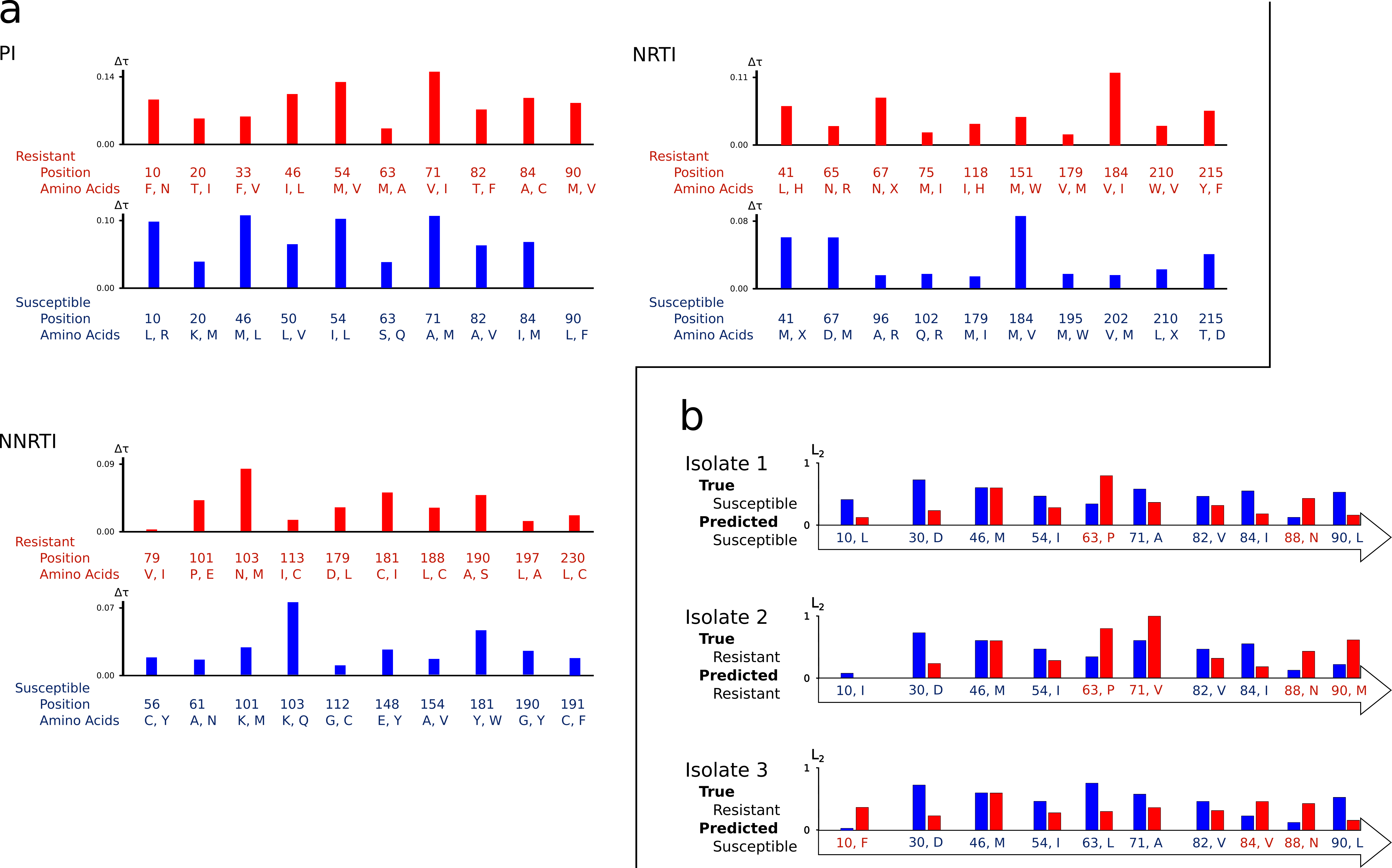}
    \caption{\textbf{a} (Global Interpretation): CMKNs can be used for data mining on biological sequences. The ten most important positions learned by the model, together with the top two contributing amino acids, are displayed. The height of the bar plot at each position indicates the normalized feature importance of that position, i.e., the mean position feature importance was subtracted from the feature importance of the specific position. Higher bars indicate more important positions. The importance of each sequence position was calculated as described in section \ref{sec:viz} and peaks were identified using a sliding window approach with a window length of 11. Afterwards, the model's learned motifs associated with the ten highest peaks were calculated (see section \ref{sec:viz}) and the two amino acids with the highest contribution to these motifs were selected. Positions displayed in red (blue) are associated with the resistant (susceptible) class. \textbf{b} (Local Interpretation): We created an exemplary visualization of CMKN's explanation capabilities. Prediction results of the nelfinavir (NFV) model for three randomly chosen input sequences are visualized by showing the learned top ten positions together with the amino acid occurring at the respective position in the input. For each position, the motif functions of the learned motifs are evaluated to identify the one with the highest $\ell_2$-norm on the input (see Section \ref{sec:viz}). If the corresponding motif is a learned resistance (susceptibility) associated motif, the position-amino-acid pair is highlighted in red (blue). The height of the bars above each position corresponds to the $\ell_2$-norm of the corresponding susceptible (blue) and resistant (red) motif functions (scaled between 0 and 1). For each isolate, the true and predicted label is displayed.}
    \label{fig:interpretability}
\end{figure*}

Apart from assessing CMKN's prediction performance, we investigated how well our models were able to learn biologically meaningful patterns from drug resistance data. For each sequence position, we calculated the position importance for each class as described in Section \ref{sec:viz} and identified peaks with a sliding window approach, i.e., the mean importance of a window of length 11 around each position was calculated and subtracted from the position importance. We selected the 10 highest peaks identified using this sliding window approach. For each peak position, the associated mean motif (of length one) as well as the two most important amino acids of this mean motif were retrieved using the approach described in Section \ref{sec:viz}. To get position importance and mean motifs for one of the three investigated drug classes (PIs, NRTIs, and NNRTIs), we averaged the importance values as well as the mean motifs over all models that belong to drugs of the same drug class (8 models for PIs, 6 models for NRTIs, and 3 models for NNRTIs). Figure \ref{fig:interpretability}\textbf{a} displays the top ten position of the resistant and susceptible class together with the top two amino acids of the corresponding mean motif for each of the three investigated drug classes. The results indicate that CMKN models are able to learn biologically meaningful patterns from real-world datasets. The most important positions identified by CMKN models correspond mainly to known drug resistance mutation (DRM) positions while the corresponding learned motifs are focused on DRMs. This result is consistent for all three tested drug types. However, CMKN models provide more than a global interpretation. Figure \ref{fig:interpretability}\textbf{b} shows the result of CMKN's local interpretation capabilities (as described in section \ref{sec:viz}) for the model trained on nelfinavir (NFV) data and three randomly selected isolates. First we identified the ten most important positions learned by the model. Afterwards, we retrieved the resistant and susceptible motifs for each position from the trained model. Using the motif functions, we were able to identify which positions the model indicated to be informative for the susceptible class and which positions were indicated to be informative for the resistant class using the procedure described in section \ref{sec:viz}. This local interpretation shows biologically meaningful patterns and can be used by domain experts to verify a prediction made by the model. For a more detailed discussion of the visualization results, see section \ref{sec:discussion}.

\subsection{Splice Site Prediction}

\begin{table}
	\centering
	\caption{Test performance on splice site benchmarks. The displayed methods include higher order Markov Chain (MC) classifiers \citep{sonnenburg2007accurate}, a combination of higher order Markov Chains and SVMs with polynomial kernel (MC-SVM) \citep{baten2006splice}, SVMs with the locality improved kernel (LIK) \citep{sonnenburg2007accurate}, SVMs with the weighted degree kernel (WD) \citep{sonnenburg2007accurate}, SVMs with the weighted degree kernel with shifts (WDS) \citep{sonnenburg2007accurate}, SpliceRover \citep{zuallaert2018splicerover}, and our CMKN. Highest numbers are shown in bold. Dashes indicate missing values in the original manuscripts.}
	\small
	\begin{tabular}{crllllrllll}
		\toprule
		\multirow{3}{*}{\textbf{Model}} & \multirow{3}{*}{\phantom{a}} & \multicolumn{4}{c}{\textbf{NN269}} & \multirow{3}{*}{\phantom{a}} & \multicolumn{4}{c}{\textbf{DGSplicer}} \\
		\cmidrule{3-6} \cmidrule{8-11}
		&& \multicolumn{2}{c}{\textbf{Acceptor}} & \multicolumn{2}{c}{\textbf{Donor}} && \multicolumn{2}{c}{\textbf{Acceptor}} & \multicolumn{2}{c}{\textbf{Donor}} \\
		&& \textbf{auROC} & \textbf{auPRC} & \textbf{auROC} & \textbf{auPRC} && \textbf{auROC} & \textbf{auPRC} & \textbf{auROC} & \textbf{auPRC} \\
		\midrule
		MC && 0.97 & 0.88 & 0.98 & 0.92 && 0.97 & 0.31 & \textbf{0.98} & 0.42 \\
		MC-SVM && 0.97 & 0.88 & 0.98 & 0.90 && 0.95 & - & 0.95 & -\\
		LIK && 0.98 & 0.92 & 0.98 & 0.93 && - & - & - & - \\
		WD && 0.98 & 0.93 & \textbf{0.99} & 0.93 && \textbf{0.98} & 0.32 & \textbf{0.98} & 0.40 \\
		WDS && \textbf{0.99} & \textbf{0.94} & 0.98 & 0.93 && 0.97 & 0.29 & 0.97 & 0.36 \\
		SpliceRover && \textbf{0.99} & - & 0.98 & - && - & - & - & - \\
		CMKN && 0.97 & \textbf{0.94} & 0.98 & \textbf{0.96} && 0.97 & \textbf{0.65} & \textbf{0.98} & \textbf{0.65} \\
		\bottomrule
	\end{tabular}
	\label{tab:splice_performance}
\end{table}

The recognition of splice sites is an important task in healthcare, since it can uncover genetic variants and differences in protein composition in individual patients. It consists of two classification problems: distinguishing decoys from true targets for acceptor sites and for donor sites.

We used two benchmarks to assess performance of our model on the splice site recognition task: NN269 \citep{reese1997improved} and DGSplicer \citep{chen2005prediction}. Both benchmarks provide test sets and are highly imbalanced. Details on training and test sets for both benchmarks can be found in the Supplement. For splice site recognition, we used the same architecture that was used for the HIV drug resistance prediction. The hyperparameter $\alpha$ was again fixed to 1. We similarly fixed the scaling parameter to $\beta = \frac{|\mathbf{x}|^2}{10}$ with $|\mathbf{x}| = 90$ for acceptor sequences and $|\mathbf{x}| = 15$ for donor sequences on the NN269 benchmark and $|\mathbf{x}| = 36$ for acceptor sequences and $|\mathbf{x}| = 18$ for donor sequences on the DGSplicer benchmark. The number of anchor points, the motif length $k$, and the positional uncertainty parameter were optimized using a grid search with 5-fold stratified cross-validation on the training data (details can be found in the Supplement). The model with the best hyperparameter combination was retrained on the whole training set and evaluated using the test set. Training success was evaluated using the area under the precision-recall curve (auPRC), to account for class imbalance, and the auROC to enable comparison with previously published models.

We compared our method to several methods that were previously applied on splice site recognition. These included higher order Markov Chain (MC) classifiers, SVMs with the locality improved kernel (LIK), the weighted degree kernel (WD), and the weighted degree kernel with shifts (WDS) published in \citep{sonnenburg2007accurate}, a method combining higher order Markov Chains and SVMs with polynomial kernel (MC-SVM) published in \citep{baten2006splice}, and a CNN architecture called SpliceRover \citep{zuallaert2018splicerover}. On the NN269 benchmark, our method performed comparable to other methods in terms of auROC and outperformed almost all competitors in terms of auPRC (see Table \ref{tab:splice_performance}). On the DGSplicer benchmark, our method performed comparable to other methods in terms of auROC, while substantially outperforming all competitors in terms of auPRC (see Table \ref{tab:splice_performance}). An evaluation of CMKN's interpretation on the splice site prediction task can be found in the Supplement.

\section{Discussion}
\label{sec:discussion}

In this work, we introduced convolutional motif kernel networks (CMKNs), a convolutional network architecture that allows for end-to-end learning within a subspace of our proposed position-aware motif kernel's RKHS.

By combining a convolutional network architecture with a kernel function, our model is able to perform robust end-to-end learning on relatively small datasets as was shown on data from Standford's HIVdb. Our model was able to generalize to validation data with only a few hundred training samples even in highly unbalanced scenarios. However, due to the fact that our model is based on a standard convolutional network architecture, CMKNs can easily be used on datasets with several hundreds of thousands of samples, as shown on the splice site prediction benchmarks. This allows to utilize our proposed kernel function on very large datasets, something that would be notoriously hard using standard kernel methods like SVMs, since the calculation of a large Gram matrix for our position-aware motif kernel is computationally very demanding.

We included accuracy and auROC as performance measures in our evaluation, since both measures are often used in the ML literature. However, on imbalanced data their informative value is decreased due to a bias towards the majority class \citep{chicco2017ten} as can be seen by considering the auROC vs. auPRC performances on the DGSplicer benchmark in Table \ref{tab:splice_performance}. Therefore, we included measures that provide better insights on imbalanced data with few positives: F1 and MCC for HIV drug resistance prediction and auPRC for splice site prediction. Considering F1, MCC, and auPRC, our model performed similar or better compared to all other models.

Another advantage of introducing kernel function evaluation into a neural architecture is the possibility to overcome the black-box nature of neural networks. Since learning within our proposed kernel layer admits a projection onto a subspace of the RKHS of our position-aware motif kernel, each output node of the kernel layer is associated with a position-motif pair. This allows for a biological interpretation of the learned weights associated with each node of the kernel layer. With these global interpretation capabilities, our model can be used as a tool for data mining on biological sequence data. We showed on HIV drug resistance data that our model is able to learn biologically meaningful patterns using standard end-to-end learning methods (see Figure \ref{fig:interpretability}\textbf{a}). The majority of the ten most important positions correspond to known DRM positions (nine for PI drugs, eight for NRTI drugs, seven for NNRTI drugs). Furthermore, the top amino acids in the learned resistant motifs reflect known DRMs while the top amino acids in the learned susceptible motifs either reflect the wildtype or none DRMs. There are three exceptions where the susceptible motif features amino acids that lead to an increased drug resistance. These exceptions are leucine (L) and valine (V) at position 50 for PI drugs, valine (V) at position 184 for NRTI drug, and aspartic acid (D) at position 215 for NRTI drugs. However, these exceptions appear to occur due to the averaging of motifs over all drugs for a specific drug class. While all of the four mentioned mutations cause an increase resistance against a subset of drugs \citep{rhee2010hiv,colonno2004identification,goudsmit1996human}, they are also a cause of increased susceptibility or have no effect for other drugs \citep{rhee2003human,bethell2012phenotypic,larder1995potential}. Valine at position 50 reduces susceptibility to fosamprenavir (FPV), lopinavir (LPV), and darunavir (DRV) but increases susceptibility to tipranavir (TPV). Leucine at position 50 confers high-level resistance to atazanavir (ATV) but increases susceptibility to all other PI drugs. For NRTI drugs, valine at position 184 reduces susceptibility to lamivudine (3TC) but increases susceptibility to zidovudine (AZT), stavudine (d4T), and tenofovir (TDF). At position 215, a mutation to aspartic acid is a so-called thymidine analog mutation that reduces susceptibility to AZT and d4T but has no effect on susceptibility to all other NRTI drugs.

Apart from the data mining capabilities of our proposed CMKN model, the motif functions enrich our model with the capability to provide local interpretations for prediction results within the data's domain. Figure \ref{fig:interpretability}\textbf{b} shows an example of the visualization capabilities of our CMKN model using nelfinavir (NFV) data, one of the PI drugs. The figure was created with the following steps. First, the trained NFV model was used to build the susceptible and resistant motifs for each of the ten most informative resistance positions learned for the NFV drug, as described in Section \ref{sec:viz} and \ref{sec:res_viz}. Afterwards, we assessed for each position if the model relates the position to the susceptible or resistant calss, as described in Section \ref{sec:viz}. For the first input, which was correctly classified as susceptible, the visualization shows that the model associated susceptible motifs with each of the positions except for position 63 and 88. However, a domain expert can quickly verify that the model falsely classified that the amino acid asparagine (N) at position 88 indicates resistance, since asparagine corresponds to the wildtype and is therefore in accordance with a susceptible isolate. Furthermore, there is no experimental evidence supporting that position 63 is associated with a resistance causing mutation. Using this knowledge, a domain expert can make an educated decision that the prediction is correct. For the correctly classified resistant input, the model associates resistant motifs with positions 63, 71, 90, and, again falsely, with position 88. Since a mutation to methionine (M) at position 90 causes a strong resistance against NFV \citep{rhee2010hiv,schapiro1996effect,craig1998hiv,zolopa1999hiv}, a domain expert could again directly validate the prediction result. The interpretation capabilities gain importance in case of a wrongly classified input as shown in the bottom part of Figure \ref{fig:interpretability}\textbf{b}. Here a domain expert would see that a susceptibility to NFV was predicted while three positions, 10, 84, and 88, are associated with resistant motifs. We again have the previously described, apparently systematic, error at position 88, but a mutation to valine (V) at position 84 causes a moderate resistance against NFV \citep{kempf2001identification}. Additionally, a mutation to phenylalanine (F) at position 10 is known to be associated with reduced in vitro susceptibility to NFV \citep{rhee2010hiv,van2009impact}. Thus, the visualization provides the domain expert with all information needed to treat the prediction outcome with the adequate caution. This shows that utilizing the proposed kernel formulation in our model's architecture, together with the proposed motif functions, can provide a visualization of a trained model's output that helps domain experts to validate the predictions.

\section{Conclusion}\label{sec:conclusion}

Our convolutional motif kernel network architecture provides inherently interpretable end-to-end learning on biological sequence data and achieves state-of-the-art performance on relevant healthcare prediction tasks, namely predicting antiretroviral drug resistance of HIV isolates and distinguishing decoys from real splice sites. 

We show that CMKN is able to learn biologically meaningful motif and position patterns on synthetic and real-world datasets. CMKN's global interpretation can foster data mining and knowledge advancement on biological sequence data. On the other hand, CMKN's local interpretation can be utilized by domain experts to judge the validity of a prediction. 

Possible future improvements include investigating a combination of different motif kernel layers to combine different motif lengths and extend the architecture to utilize meaningful combinations of motifs. Another improvement that we want to explore in future work is the extension of the kernel formulation to multi-layer networks while securing the interpretation capabilities.

\section*{Data and Code Availability}
Source code, pre-processing scripts, and experimental scripts are available on GitHub \href{https://github.com/jditz/CMKN}{here}. HIV drug resistance data can be found online on \href{https://hivdb.stanford.edu/pages/genopheno.dataset.html}{HIVdb}. The NN269 benchmark can be downloaded from the \href{https://cs.gmu.edu/~ashehu/sites/default/files/tools/EFFECT_2013/data.html}{George Manson University}. The DGSplicer benchmark was provided by the original author, \href{https://icannwiki.org/Chung-Chin_Lu}{Prof. Chung-Chin Lu}.

\begin{ack}
The authors would like to thank Prof. Chung-Chin Lu for providing the DGSplicer benchmark. Funded by the Deutsche Forschungsgemeinschaft (DFG, German Research Foundation) under Germany’s Excellence Strategy – EXC number 2064/1 – Project number 390727645. This research was supported by the German Federal Ministry of Education and Research (BMBF) project 'Training Center Machine Learning, T\"{u}bingen' with grant number 01|S17054. This work was supported by the German Federal Ministry of Education and Research (BMBF): T\"{u}bingen AI Center, FKZ: 01IS18039A.
\end{ack}

\bibliographystyle{apalike}
\bibliography{references}

\newpage

\appendix

\section{Societal and Environmental Impact}
Medical data are notoriously biased against minorities and there are numerous examples for machine learning models that learn this biases and have a severe deterioration in performance with regard to minorities (see e.g., \citep{morley2020ethics}). We did not include a statement on the societal impact of our work into the main manuscript due to the used benchmarks. The HIV benchmark consists of viral sequences, which renders the evaluation regarding human metadata unnecessary. On the contrary, improving the treatment of HIV infection and AIDS has a beneficial impact on society. On the other hand, the splice site benchmarks are created using human genes. However, the original authors did not include any necessary metadata into the benchmarks, which renders a evaluation as described above impossible. We will include the evaluation of potential bias against minorities in future work, if the model is used in a real-world application on medical data that is potentially biased. Since this bias normally arises from the data, such considerations are task-specific and not model-specific.

All experiments were conducted using a single NVIDIA GeForce GTX 1080 Ti GPU. All experiments together required a total of 124 hours of computing time. This resulted in total emissions of 13.39 kg CO$_2$e, which is equivalent to burning 6.71 kg of coal. To compensate this emissions, 0.22 tree seedlings have to sequester carbon for 10 years. These estimations were calculated using the Machine Learning Impact calculator\footnote{https://mlco2.github.io/impact/} by Lacoste and colleagues (Alexandre Lacoste et al.: \textit{Quantifying the carbon emissions of machine learning.} arXiv preprint arXiv:1910.09700 (2019)).

\section{Setting the Scaling Parameter}\label{sec:AppBeta}
\begin{figure}[!h]
	\centering
	\includegraphics[width=\textwidth]{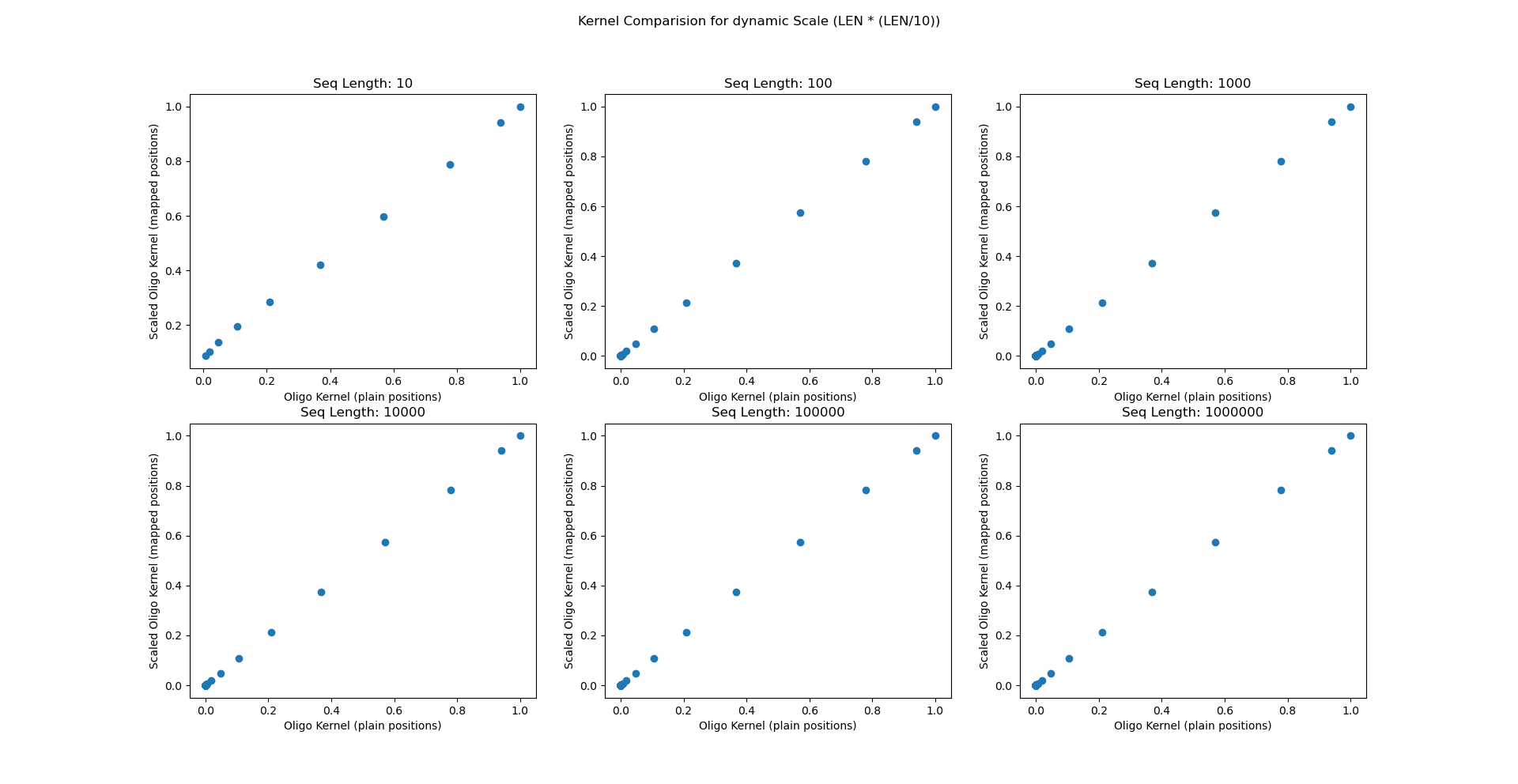}
	\caption{Comparison of the exponential term of the oligo kernel with the position kernel as introduced in the main manuscript in Eq. (1). Six different sequence lengths were tested: 10, 100, 1000, 10 000, 100 000, and 1 000 000. The scaling parameter $\beta$ of the position kernel was set to $\frac{|\mathbf{x}|^2}{10}$, where $|\mathbf{x}|$ denotes the corresponding sequence length. As noticeable in the plots, the chosen parameter value compensates for the reduced absolute distances of sequence positions and their dependence on the sequence length due to the projection onto the unit circle introduced in the main manuscript.}
	\label{fig:scaleParam}
\end{figure}

Given a sequence $\mathbf{x}$ with length $|\mathbf{x}|$, we found empirically that setting $\beta = \frac{|\mathbf{x}|^2}{10}$ compensates for the transformation of sequence positions as introduced in the manuscript (see Figure \ref{fig:scaleParam}).

\section{Normalized Position Frequency Matrix}\label{sec:nPWM}
A set of biological sequences of length $k$ can be easily transformed in to a position frequency matrix (PFM) by counting the occurrences of each nucleotide or amino acid from an alphabet $\mathbf{A}$ at each position and constructing a matrix $M=(m_{ij})_{i=1, \ldots,|\mathbf{A}|; j=1,\ldots,k}$, where $m_{ij}$ is equal to the number of occurrences of nucleotide or amino acid $i$ at position $j$. The PFM can then be easily transformed into a normalized position frequency matrix (nPFM) $M_{\textrm{norm}}=(m_{\textrm{norm}, ij})_{i=1,\ldots,|\mathbf{A}|; j=1,\ldots,k}$ by dividing each entry of every column $m_j=\sum_i m_{ij}$ of $M$ by the $\ell_2$-norm of the respective column, i.e.
\begin{equation}
\label{eqn:nPFM}
m_{\textrm{norm}, ij} = \frac{m_{ij}}{\norm{m_j}_2} .
\end{equation} 
Using this definition of an nPFM, the set of all $k$-mers is a subset of the set of all sequence motifs. Considering a k-mer as a special motifs, each column of the corresponding nPFM has only one element set to one while all other elements are set to zero.

\section{Orthogonal Projection onto the Subspace $\mathcal{E}$}\label{sec:AppDeri}
We denote a motif position pair by $\mathbf{y} = (\omega, p)$. Then, the explicit parametrization derived by Mairal in \citep{mairal2016end} Appendix A also holds for our kernel. This can be easily shown by simple calculus using Mairal's definition:
\begin{equation}
f_{\mathbf{y}} := \sum_{j = 1}^{n} \alpha_j^* \varphi(z_j) \qquad \text{with} \qquad \alpha^* \in \argmin_{\alpha \in \mathbb{R}^{n}} \norm{\varphi(\mathbf{y}) - \sum_{j = 1}^{n} \alpha_j \varphi(z_j)}_\mathcal{H}^2 .
\end{equation}
Then the following holds:
\begin{align}
\norm{\varphi(\mathbf{y}) - \sum_{j = 1}^{n} \alpha_j \varphi(z_j)}_\mathcal{H}^2 &= \langle \varphi(\mathbf{y}), \varphi(\mathbf{y}) \rangle_\mathcal{H} - 2 \langle \varphi(\mathbf{y}), \sum_{j = 1}^{n} \alpha_j \varphi(z_j) \rangle_\mathcal{H} + \langle \sum_{j = 1}^{n} \alpha_j \varphi(z_j), \sum_{i = 1}^{n} \alpha_i \varphi(z_i) \rangle_\mathcal{H} \nonumber \\
&= 1 - 2 \sum_{j = 1}^{n} \langle \varphi(\mathbf{y}), \alpha_j \varphi(z_j) \rangle_\mathcal{H} + \sum_{j = 1}^{n} \sum_{i = 1}^{n} \langle \alpha_j \varphi(z_j), \alpha_i \varphi(z_i) \rangle_\mathcal{H} \nonumber \\
&= 1 - 2 \sum_{j = 1}^{n} \alpha_j K_0(\mathbf{y}, z_j) + \sum_{j = 1}^{n} \sum_{i = 1}^{n} \alpha_j \alpha_i K_0(z_j, z_i) \nonumber \\
&= 1 - 2 \alpha^T K_{Z}(\mathbf{y}) + \alpha^T K_{ZZ} \alpha
\end{align}
Here, $K_{ZZ}=(K_0(z_i, z_j))_{i=1,\ldots,n;j=1,\ldots,n}$ is the Gram matrix formed over the anchor points $z_1, \ldots, z_n$, $K_{ZZ}^{-\frac{1}{2}}$ is the (pseudo)-inverse square root of the Gram matrix, and $K_{Z}((\omega, p)) = (K_0(z_1, (\omega, p)), \ldots, K_0(z_n, (\omega, p)))^T$. $K_0$ is the kernel function introduced in the main manuscript. From here the argument is similar to Mairal's argument in the Appendix of \citep{mairal2016end}.

\section{Deriving the Position-Aware Motif Kernel from Motif Functions}\label{sec:PAMKfromMF}
For a sequence $\mathbf{x}$ over an alphabet $\mathbf{A}$ and a motif $\chi \in \mathbb{R}^{|\mathbf{A}|k}$, we define the motif function as
\begin{equation}
\label{eqn:motifFunction}
\phi_{\mathbf{x}}(\chi, t) = \sum^{|\mathbf{x}|}_{p=1} \phi_{\mathbf{x}, p}(\chi, t),
\end{equation}
with 
\begin{equation*}
\phi_{\mathbf{x}, p}(\chi, t) = \exp \left( -\alpha \norm{\chi - \omega_{p}}_2^2 - \frac{\beta}{2\sigma^2} \norm{t - \tilde{p}}_2^2 \right)
\end{equation*}
and position variable $t \in \mathbb{R}^{2}$.

Consider two sequences $\mathbf{x}$ and $\mathbf{x}'$ of same length, i.e. $|\mathbf{x}|=|\mathbf{x}'|$. The motif kernel between them is given as the inner product of the associated motif functions $\phi_{\mathbf{x}}$ and $\phi_{\mathbf{x}'}$: 
\begin{equation}
\label{eqn:int1}
\begin{split}
K_{\textrm{PAM}}(\mathbf{x}, \mathbf{x}')
=
\phi_{\mathbf{x}} \cdot \phi_{\mathbf{x}'}
=
\int \int \phi_{\mathbf{x}}(\chi, t) \cdot \phi_{\mathbf{x'}}(\chi, t) \mathop{d\chi} \mathop{dt} \\
=
\sum_{p = 1}^{|\mathbf{x}|} \sum_{q = 1}^{|\mathbf{x}'|} \int \int e^{ -\alpha \norm{\chi - \omega_{p}}_2^2 - \frac{\beta}{2\sigma^2} \norm{t - \tilde{p}}_2^2 } \cdot e^{ -\alpha \norm{\chi - \omega_{q}}_2^2 - \frac{\beta}{2\sigma^2} \norm{t - \tilde{q}}_2^2 } \mathop{d\chi} \mathop{dt} \\
=
\sum_{p = 1}^{|\mathbf{x}|} \sum_{q = 1}^{|\mathbf{x}'|} \int \int e^{ -\alpha \norm{\chi - \omega_{p}}_2^2 } \cdot e^{ -\alpha \norm{\chi - \omega_{q}}_2^2 } \cdot e^{ -\frac{\beta}{2\sigma^2} \norm{t - \tilde{p}}_2^2 } \cdot e^{ -\frac{\beta}{2\sigma^2} \norm{t - \tilde{q}}_2^2 } \mathop{d\chi} \mathop{dt} \\
=
\sum_{p = 1}^{|\mathbf{x}|} \sum_{q = 1}^{|\mathbf{x}'|} \int e^{ -\alpha \left( \norm{\chi - \omega_{p}}_2^2 + \norm{\chi - \omega_{q}}_2^2 \right) } \mathop{d\chi} \cdot \int e^{ -\frac{\beta}{2\sigma^2} \left( \norm{t - \tilde{p}}_2^2 + \norm{t - \tilde{q}}_2^2 \right) } \mathop{dt} \\
\stackrel{(\ref{eqn:int2})}{=}
\sum_{p = 1}^{|\mathbf{x}|} \sum_{q = 1}^{|\mathbf{x'}|} \sqrt{\frac{\pi}{2\alpha}} e^{-\frac{\alpha}{2}\norm{\omega_{p} - \omega_{q}}_2^2} \sqrt{\frac{\pi\sigma^2}{\beta}} \cdot e^{-\frac{\beta}{4\sigma^2}\norm{\tilde{p} - \tilde{q}}_2^2} \\
\stackrel{(\ref{eqn:linearization})}{=}
\sqrt{\frac{\pi^2\sigma^2}{2\alpha\beta}} \sum_{p = 1}^{|\mathbf{x}|} \sum_{q = 1}^{|\mathbf{x}'|} e^{\alpha\left(\omega_{p}^{T}\omega_{q} - k \right) + \frac{\beta}{2\sigma^2}\left(\tilde{p}^{T}\tilde{q} - 1\right)},
\end{split}
\end{equation}
with $\omega_{p}, \omega_{q} \in \mathbb{R}_{+}^{|\mathbf{A}|k}$ and $\tilde{p}, \tilde{q}$ from the upper half of the unit circle.
The above derivation uses the fact that the variables $\chi \in \mathbb{R}^{|\mathbf{A}|k}$ and $t \in \mathbb{R}^{2}$ are not restricted and thus (using the substitutions $y=\chi$, $\gamma=\alpha$, $\delta=\omega_{p}$, $\epsilon=\omega_{q}$, $N=|\mathbf{A}|k$ and $y=t$, $\gamma=\frac{\beta}{2\sigma^2}$, $\delta=\tilde{p}$, $\epsilon=\tilde{q}$, $N=2$, respectively) can be removed via
\begin{equation}
\label{eqn:int2}
\begin{split}
\int e^{ -\gamma \left( \norm{y - \delta}_2^2 + \norm{y - \epsilon}_2^2 \right) } \mathop{dy} \\
=
\int_{-\infty}^{\infty} \ldots \int_{-\infty}^{\infty} e^{ -\gamma \left( (y_1 - \delta_1)^2 + \ldots + (y_N - \delta_N)^2 + (y_1 - \epsilon_1)^2 + \ldots + (y_N - \epsilon_N)^2 \right) } \mathop{dy_1} \ldots \mathop{dy_N} \\
=
\int_{-\infty}^{\infty} e^{ -\gamma (y_1 - \delta_1)^2} e^{ -\gamma (y_1 - \epsilon_1)^2 } \mathop{dy_1} \cdot \ldots \cdot \int_{-\infty}^{\infty} e^{ -\gamma (y_N - \delta_N)^2} e^{ -\gamma (y_N - \epsilon_N)^2 } \mathop{dy_N} \\
=
\int_{-\infty}^{\infty} e^{ -\gamma \left( (y_1 - \delta_1)^2 + (y_1 - \epsilon_1)^2 \right) } \mathop{dy_1} \cdot \ldots \cdot \int_{-\infty}^{\infty} e^{ -\gamma \left( (y_N - \delta_N)^2 + (y_N - \epsilon_N)^2 \right) } \mathop{dy_N} \\
=
\int_{-\infty}^{\infty} e^{ -2\gamma y_1^2 + 2\gamma(\delta_1 + \epsilon_1)y_1 - \gamma(\delta_1^2 + \epsilon_1^2) } \mathop{dy_1} \cdot \ldots \cdot \int_{-\infty}^{\infty} e^{ -2\gamma y_N^2 + 2\gamma(\delta_N + \epsilon_N)y_N - \gamma(\delta_N^2 + \epsilon_N^2)  } \mathop{dy_N} \\
\stackrel{(\ref{eqn:int3})}{=}
\sqrt{\frac{\pi}{2\gamma}} e^{ \frac{\left(2\gamma(\delta_1 + \epsilon_1)\right)^2}{8\gamma} - \gamma(\delta_1^2 + \epsilon_1^2) } \cdot \ldots \cdot \sqrt{\frac{\pi}{2\gamma}} e^{ \frac{\left(2\gamma(\delta_N + \epsilon_N)\right)^2}{8\gamma} - \gamma(\delta_N^2 + \epsilon_N^2) } \\
=
\sqrt{\frac{\pi}{2\gamma}} \left( e^{ \gamma \left( \frac{1}{2}(\delta_1^2 + 2\delta_1\epsilon_1 + \epsilon_1^2) - \delta_1^2 - \epsilon_1^2 \right) } \cdot \ldots \cdot e^{ \gamma \left( \frac{1}{2}(\delta_N^2 + 2\delta_N\epsilon_N + \epsilon_N^2) - \delta_N^2 - \epsilon_N^2 \right) } \right) \\
=
\sqrt{\frac{\pi}{2\gamma}} \left( e^{ -\frac{\gamma}{2} \left( \delta_1^2 - 2\delta_1\epsilon_1 + \epsilon_1^2 \right) } \cdot \ldots \cdot e^{ -\frac{\gamma}{2} \left( \delta_N^2 - 2\delta_N\epsilon_N + \epsilon_N^2 \right) } \right) \\
=
\sqrt{\frac{\pi}{2\gamma}} \left( e^{ -\frac{\gamma}{2} (\delta_1 - \epsilon_1)^2 } \cdot \ldots \cdot e^{ -\frac{\gamma}{2} (\delta_N - \epsilon_N)^2 } \right) \\
=
\sqrt{\frac{\pi}{2\gamma}} e^{ -\frac{\gamma}{2} \norm{\delta - \epsilon}_2^2}
\end{split}
\end{equation}
using
\begin{equation}
\label{eqn:int3}
\int_{-\infty}^{\infty} e^{ -ay^2 + by + c } \mathop{dy} =
\sqrt{\frac{\pi}{a}} e^{\frac{b^2}{4a} + c}
\end{equation}
with $a = 2\gamma$, $b = 2\gamma(\delta_i + \epsilon_i)$, and $c = - \gamma(\delta_i^2 + \epsilon_i^2)$.

The last equality in Eq. (\ref{eqn:int1}) results from the normalisation of the nPFM as defined in Eq. (\ref{eqn:nPFM}), i.e., the column-wise unit $\ell_2$-norm demanded earlier in Section 5.1 of the main manuscript. Consider two normalized position frequency matrices (nPFM) $A \in \mathbb{R}_{+}^{|\mathbf{A}| \times k}$ and $B \in \mathbb{R}_{+}^{|\mathbf{A}| \times k}$, with $|\mathbf{A}|$ being the size of the alphabet over which the motif is created and $k$ being the length of the motif. One can define two vectors $a$ and $b$ given as flattened nPFMs $A$ and $B$, i.e., the columns are concatenated to convert the matrices into vectors. Utilizing that each column of a nPFM has unit $\ell_2$-norm by definition, the following equality is obtained:
\begin{equation}
\label{eqn:linearization}
\begin{split}
\norm{a - b}^2 = (a_{11} - b_{11})^2 + ... + (a_{|A|1} - b_{|A|1})^2 + ... + (a_{1k} - b_{1k})^2 + ... + (a_{|A|k} - b_{|A|k})^2 \\
=
\underbrace{a_{11}^2 + ... + a_{|A|1}^2}_{=1} + \underbrace{b_{11}^2 + ... + b_{|A|1}^2}_{=1} + ... + \underbrace{a_{1k}^2 + ... + a_{|A|k}^2}_{=1} + \underbrace{b_{1k}^2 + ... + b_{|A|k}^2}_{=1} \\
- \underbrace{2( a_{11}b_{11} + ... + a_{|A|1}b_{|A|1} + ... + a_{1k}b_{1k} + ... + a_{|A|k}b_{|A|k} )}_{=ab} \\
=
2k - 2ab \quad \Rightarrow  \quad -\frac{1}{2}\norm{a - b}^2 = ab - k
\end{split}
\end{equation}

\section{HIVdb}\label{sec:AppHIVdb}

\begin{sidewaystable}
	\ra{1.3}
	\centering
	\caption{Overview of the HIVdb datasets. For each drug, basic dataset statistics are displayed as well as the best values for the positional uncertainty parameter $\sigma$ and the number of anchor points. Furthermore, mean values as well as standard deviations for four performance parameters (accuracy, F1 score, area under the receiver operating characteristic curve (auROC), and Matthew's correlation coefficient (MCC)) achieved on validation sets during a stratified 5-fold cross validation are shown.}
	\hspace{-1cm}\begin{tabular}{@{}cccccrccrllll@{}}
		\toprule
		\textbf{Type} & \textbf{Drug} & \textbf{Samples} & \multicolumn{2}{c}{\textbf{Class Distribution (\%)}} & \phantom{abc} & \multicolumn{2}{c}{\textbf{Best Parameters}} & \phantom{abc} & \textbf{Accuracy} & \textbf{F1 Score} & \textbf{auROC} & \textbf{MCC} \\
		\cmidrule{4-5} \cmidrule{7-8}
		& & & susceptible & resistant && $\sigma$ & anchors && & & & \\
		\midrule
		PI\\
		& ATV & 477 & 59.3 & 40.7 && 16 &  50 && 0.918 $\pm$ 0.041 & 0.901 $\pm$ 0.051 & 0.969 $\pm$ 0.011 & 0.832 $\pm$ 0.086 \\
		& DRV & 273 & 87.9 & 12.1 &&  4 &  99 && 0.945 $\pm$ 0.033 & 0.779 $\pm$ 0.118 & 0.970 $\pm$ 0.019 & 0.757 $\pm$ 0.132 \\
		& FPV & 737 & 65.3 & 34.7 &&  8 &  99 && 0.912 $\pm$ 0.022 & 0.873 $\pm$ 0.032 & 0.965 $\pm$ 0.018 & 0.806 $\pm$ 0.049 \\
		& IDV & 771 & 58.5 & 41.5 && 16 &  75 && 0.930 $\pm$ 0.019 & 0.915 $\pm$ 0.022 & 0.975 $\pm$ 0.008 & 0.856 $\pm$ 0.038 \\
		& LPV & 612 & 67.2 & 32.8 &&  4 &  50 && 0.920 $\pm$ 0.012 & 0.881 $\pm$ 0.017 & 0.970 $\pm$ 0.006 & 0.822 $\pm$ 0.026 \\
		& NFV & 793 & 46.5 & 53.5 && 16 &  99 && 0.927 $\pm$ 0.014 & 0.932 $\pm$ 0.013 & 0.971 $\pm$ 0.008 & 0.853 $\pm$ 0.027 \\
		& SQV & 775 & 63.4 & 36.6 &&  8 &  99 && 0.939 $\pm$ 0.015 & 0.916 $\pm$ 0.022 & 0.980 $\pm$ 0.009 & 0.869 $\pm$ 0.033 \\
		& TPV & 306 & 76.8 & 23.2 &&  8 &  75 && 0.876 $\pm$ 0.044 & 0.725 $\pm$ 0.108 & 0.898 $\pm$ 0.032 & 0.646 $\pm$ 0.135 \\
		NRTI\\
		& 3TC & 477 & 45.3 & 54.7 && 16 & 120 && 0.937 $\pm$ 0.023 & 0.944 $\pm$ 0.023 & 0.973 $\pm$ 0.013 & 0.879 $\pm$ 0.049 \\
		& ABC & 480 & 36.9 & 63.1 && 16 & 180 && 0.919 $\pm$ 0.017 & 0.935 $\pm$ 0.014 & 0.962 $\pm$ 0.017 & 0.828 $\pm$ 0.033\\
		& AZT & 477 & 58.5 & 41.5 && 16 & 120 && 0.899 $\pm$ 0.025 & 0.882 $\pm$ 0.026 & 0.965 $\pm$ 0.016 & 0.799 $\pm$ 0.047 \\
		& D4T & 479 & 57.6 & 42.4 && 16 & 120 && 0.854 $\pm$ 0.045 & 0.826 $\pm$ 0.056 & 0.916 $\pm$ 0.030 & 0.701 $\pm$ 0.094 \\
		& DDI & 479 & 52.0 & 48.0 && 16 & 240 && 0.862 $\pm$ 0.033 & 0.854 $\pm$ 0.038 & 0.914 $\pm$ 0.029 & 0.726 $\pm$ 0.066 \\
		& TDF & 395 & 73.4 & 26.6 &&  8 & 240 && 0.843 $\pm$ 0.034 & 0.711 $\pm$ 0.070 & 0.871 $\pm$ 0.050 & 0.609 $\pm$ 0.093 \\
		NNRTI\\
		& EFV & 511 & 55.4 & 44.6 && 16 & 180 && 0.924 $\pm$ 0.019 & 0.913 $\pm$ 0.023 & 0.965 $\pm$ 0.015 & 0.846 $\pm$ 0.039 \\
		& ETR & 183 & 66.1 & 33.9 &&  8 & 120 && 0.886 $\pm$ 0.043 & 0.829 $\pm$ 0.065 & 0.909 $\pm$ 0.061 & 0.750 $\pm$ 0.093 \\
		& NVP & 515 & 47.0 & 53.0 && 16 & 120 && 0.920 $\pm$ 0.019 & 0.924 $\pm$ 0.019 & 0.966 $\pm$ 0.010 & 0.843 $\pm$ 0.036 \\
		& RPV &  83 & 66.3 & 33.7 &&  \multicolumn{7}{c}{\textit{Removed from analysis due to low sample count}} \\
		\bottomrule
	\end{tabular}
	\label{tab:hivdb_stats}
\end{sidewaystable}

HIVdb is one of the largest, publicly available databases containing resistance information against antiretroviral drugs that are used for the treatment of an HIV infection or an acquired immunodeficiency syndrome (AIDS). We used available information for eight PI drugs, six NRTI drugs, and four NNRTI drugs (data was downloaded on 14 April 2021). The PI dataset contained 819 isolates, each represented by an amino acid sequence of length 99. The NRTI dataset contained 489 isolates, each represented by an amino acid sequence of length 240. Finally, the NNRTI dataset contained 583 isolates, each represented by an amino acid sequence of length 240. For each isolate, available drug resistance information for each drug of the corresponding type was given as a change in fold resistance compared to the wildtype. We used thresholds provided by HIVdb to convert fold resistance values into the discrete classes 'susceptible' and 'resistant'.

HIVdb stores the isolate information as a tab-separated table. Each position is represented either by a '-', if the amino acid is the same as in the wild type, or by the mutation at that position. Furthermore, there are columns for each drug with the fold resistance of the isolate compared to the wild type. We provide a script that translates these tables into a FASTA file that can be used directly as input for the provided \textit{DataSet} object. HIVdb provides two fold resistance thresholds for each of the drugs within the database. These thresholds are used by our preparation script to assign each isolate to either the 'low resistance', 'medium resistance', or 'high resistance' class. The \textit{DataSet} object combines 'medium resistance' and 'high resistance' classes into the 'resistant' class on data access.

Table \ref{tab:hivdb_stats} provides detailed information about sample count, class distribution, best performing parameters, and classification performances for each of the tested drugs. The amount of isolates for each of the drugs differs from the total number of isolates, since not all isolates had fold resistance information for each of the drugs. We excluded drugs with less than 100 isolates from all further evaluations. Each of the drugs is indicated using the official three-letter abbreviation. The corresponding complete names and trade names are written below.

\begin{center}
\begin{tabular}{llll}
	\toprule
	\textbf{Type} & \textbf{Abbreviation} & \textbf{Name} & \textbf{Trade Name} \\
	\midrule
	PI\\
	& ATV & Atazanavir & Reyataz, Evotaz, others \\
	& DRV & Darunavir & Prezista, Prezcobix, others \\
	& FPV & Fosamprenavir & Lexiva, Telzir \\
	& IDV & Indinavir & Crixivan \\
	& LPV & Lopinavir & - \\
	& NFV & Nelfinavir & Viracept \\
	& SQV & Saquinavir & Invirase, Fortovase \\
	& TPV & Tipranavir & Aptivus \\
	NRTI\\
	& 3TC & Lamivudine & Epivir, Epivir-HBV, Zeffix, others \\
	& ABC & Abacavir & Ziagen, others \\
	& AZT\footnote{Azidothymidine is also known as Zidovudine (ZDV)} & Azidothymidine & Retrovir, others \\
	& D4T & Stavudine & Zerit \\
	& DDI & Didanosine & Videx \\
	& TDF & Tenofovir disoproxil & Viread, others \\
	NNRTI\\
	& EFV & Efavirenz & Sustiva, Stocrin, others \\
	& ETR & Etravirine & Intelence \\
	& NVP & Nevirapine & Viramune \\
	& RPV & Rilpivirine & Edurant, Rekambys \\
	\bottomrule
\end{tabular}
\end{center}

\section{Splice Site Benchmarks}\label{sec:AppSplice}
\begin{table}[!h]
    \centering
    \caption{Overview of the benchmarks used to evaluate CMKN on the splice site prediction task.}
    \begin{tabular}{ccrcccrccc}
    \toprule
    \textbf{Benchmark} & \textbf{Type} & \phantom{abc} & \textbf{Samples} & \multicolumn{2}{c}{\textbf{Class Distribution (\%)}} & \phantom{abc} & \multicolumn{3}{c}{\textbf{Best Parameters}} \\
    \cmidrule{5-6} \cmidrule{8-10}
     & & && pos & neg && $k$ & $\sigma$ & anchors \\
    \midrule
    NN269 \\
     & Acceptor (train) && 5788 & 19.3 & 80.7 && \multirow{2}{*}{2} & \multirow{2}{*}{16} & \multirow{2}{*}{45} \\
     & Acceptor (test) && 1089 & 19.1 & 80.9 && \\
     & Donor (train) && 5256 & 21.2 & 78.8 && \multirow{2}{*}{5} & \multirow{2}{*}{4} & \multirow{2}{*}{30} \\
     & Donor (test) && 990 & 21.0 & 79.0 && \\
    DGSplicer \\
     & Acceptor (train) && 322155 & 0.6 & 99.4 && \multirow{2}{*}{5} & \multirow{2}{*}{8} & \multirow{2}{*}{45} \\
     & Acceptor (test) && 80539 & 0.6 & 99.4 && \\
     & Donor (train) && 228267 & 0.8 & 99.2 && \multirow{2}{*}{5} & \multirow{2}{*}{4} & \multirow{2}{*}{45} \\
     & Donor (test) && 57067 & 0.8 & 99.2 && \\
    \bottomrule
    \end{tabular}
    \label{tab:splice_bench}
\end{table}

The NN269 benchmark consists of 1324 true targets for both acceptor and donor datasets that were collected from 269 human genes. The acceptor sequences consist of 90 nucleotides with the splice site acceptor dimer AG at positions 69 to 70. The donor sequences consist of 15 nucleotides with the splice site donor dimer GT at positions 8 to 9. Furthermore, the benchmark includes 5553 acceptor decoys and 4922 donor decoys. Each decoy sequence has the splice site dimers (AG for acceptor sites, GT for donor sites) at the same positions as the true targets. The DGSplicer benchmark consists of 2380 true acceptor targets and 2379 true donor targets that were extracted from 462 multi-exon human genes. The acceptor sequences have a length of 36 nucleotides with the dimer AG at positions 26 to 27. The donor sequences have a length of 18 nucleotides with the dimer GT at positions 10 to 11. As with NN269, decoy sequences are included in the benchmark, i.e., 400314 pseudo acceptor sites and 282955 pseudo donor sites. Both benchmarks are split into training and test sets. The detailed statistics about the used benchmarks for the prediction of splice sites can be found in Table \ref{tab:splice_bench}. To deal with the huge imbalance between positive and negative samples within the DGSplicer benchmark, we randomly under-sampled the negative samples in the training sets to achieve a class ratio of $\frac{N_p}{N_{rn}} = 0.25$, where $N_p$ is the number of samples in the positive class and $N_{rn}$ is the number of samples in the negative class after resampling. Class ratios in validation and test sets were kept unchanged.

\section{Hyperparameter Optimization}\label{sec:AppHypOp}
Experiments were conducted with Python 3.6.9 using imbalanced-learn 0.7.0, numpy 1.18.5, scikit-learn 0.23.2, scipy 1.5.2, and torch 1.6.0.

\subsection{HIV Drug Resistance Prediction}
\subsubsection{CMKN}
CMKN models have 5 hyperparameters associated with the motif kernel layer that need to be optimized. These include the k-mer length $k$, the motif comparison parameter $\alpha$, the position scaling parameter $\beta$, the positional uncertainty parameter $\sigma$, and the number of anchor points. 

We used prior (biological) knowledge to fix three of the hyperparameters. This allowed us to reduce computation time and lower the $\mathrm{CO}_2$ footprint of our experiments. We fixed the motif length $k$ to $1$, since considering longer motifs offers no advantage in the case of resistance mutations against antiretroviral drugs. Previously conducted research suggest that a drug resistance is caused by exchanging single amino acids which justifies to set the motif length to $1$. Additionally, we fixed $\alpha$ to $1.0$ to ensure that the impact of inexact motif matching is \textit{not} reduced. This takes the fact into account that some positions can have several different mutations causing resistance against antiretroviral drugs and the exchange of a single amino acid can cause a drug resistance. Therefore, inexact motif matching contains valuable information which justifies the value for $\alpha$ we used. Lastly, the position scaling parameter $\beta$ was fixed to $\frac{|\mathbf{x}|^2}{10}$, where $|\mathbf{x}|$ denotes the length of the input sequences. This value compensates for the transformation of the sequence positions as described in Section \ref{sec:AppBeta}.

The other two hyperparameters, the positional uncertainty $\sigma$ and the number of anchor points, were optimized using a simple grid search. We used a small but sensible number of values for the grid to keep the $\mathrm{CO}_2$ footprint of our experiments as low as possible. The following values were used for the grid search:
\begin{align}
\sigma &\in \{ 1, 2, 4, 8, 16 \} \nonumber \\
\#\textrm{anchors}\,_{\textrm{PI}} &\in \{ 50, 75, 99 \} \nonumber \\
\#\textrm{anchors}\,_{\textrm{NRTI}} &\in \{ 120, 180, 240 \} \nonumber \\
\#\textrm{anchors}\,_{\textrm{NNRTI}} &\in \{ 120, 180, 240 \} \nonumber 
\end{align}
The different choices for the number of anchors for PI and NRTI/NNRTI drugs were chosen to take account of the different sequence lengths. The optimal parameter choice for each of the drugs can be found in Table \ref{tab:hivdb_stats}.

All CMKN models were trained using PyTorch's implementation of the ADAM algorithm and class-balanced loss as introduced in \citep{cui2019class}. Training lasted for 200 epochs and no early-stopping was used. The learning rate started at $0.1$ and was dynamically adjusted using the \textit{ReduceLROnPlateau} method.

The value combination that maximized the most of the four performance measures, accuracy, F1 score, auROC, and MCC, was selected for further analysis.

\subsubsection{Models used for Comparison}
We used standard parameters for the random forest classifiers trained on our HIVdb datasets that matched the parameters used in \citep{raposo2020random}. The number of trees was set to 500 while the number of features to look at, when searching for the best split, was set to the square root of the total number of features. For all other parameters the default values of the \textit{sklearn.ensemble.RandomForestClassifier} class were used. This is coherent with the training procedure used in \citep{raposo2020random}. The input sequences were ordinal encoded before passing them to the classifier, i.e., each letter was replaced by an integer reflecting the position of the letter in the alphabet.

For support vector machines (SVMs) with polynomial kernel, we performed a grid search to optimize the degree of the polynomial kernel and the regularization parameter. The grid consisted of the following values:
\begin{align}
\textrm{C} &\in \{ 10^{-5}, 10^{-4}, 10^{-3}, 10^{-2}, 10^{-1}, 1, 10 \} \nonumber \\
\textrm{degree} &\in \{ 1, 2, 3, 4, 5 \} \nonumber
\end{align}
Similar to the CMKN models, the parameter combination that maximized the most of the four performance measures, accuracy, F1 score, auROC, and MCC, was selected for further analysis. We used the same ordinal encoding that was used in the RF experiment to encode the input for the SVMs. For the model selection of SVMs with oligo kernel, we used the input encoding as described in \citep{meinicke2004oligo}. The hyperparameters of this method are the length of $k$-mers denoted by $k$, the position uncertainty parameter $\sigma$, and the regularization parameter C. Similar to CMKNs, $k$ was fixed to 1 due to biological reasons. The other two hyperparameter were optimized with the following grid:
\begin{align}
\textrm{C} &\in \{ 10^{-5}, 10^{-4}, 10^{-3}, 10^{-2}, 10^{-1}, 1, 10 \} \nonumber \\
\sigma &\in \{ 1, 2, 4, 8, 16 \} \nonumber
\end{align}
The optimal parameters for the SVM models were:

\begin{center}
\begin{tabular}{crccrcc}
    \toprule
    \textbf{Drug} & \phantom{abc} & \multicolumn{2}{c}{$\textrm{SVM}_\textrm{poly}$} & \phantom{abc} & \multicolumn{2}{c}{$\textrm{SVM}_\textrm{oligo}$} \\
    \cmidrule{3-4} \cmidrule{6-7}
     && C & degree && C & $\sigma$ \\
    \midrule
    ATV && 1.0 & 3 && 0.1 & 4 \\
    DRV && 0.001 & 1 && 0.01 & 1 \\
    FPV && 0.1 & 5 && 1.0 & 1 \\
    IDV && 10.0 & 2 && 1.0 & 1 \\
    LPV && 1.0 & 3 && 1.0 & 1 \\
    NFV && 10.0 & 2 && 1.0 & 1 \\
    SQV && 1.0 & 3 && 1.0 & 1 \\
    TPV && 0.001 & 2 && 0.1 & 1 \\
     \\
    3TC && 0.1 & 4 && 0.1 & 1 \\
    ABC && 0.1 & 4 && 0.1 & 1 \\
    AZT && 0.1 & 5 && 1.0 & 1 \\
    D4T && 0.1 & 3 && 0.1 & 1 \\
    DDI && 0.1 & 3 && 0.1 & 1 \\
    TDF && 0.01 & 5 && 1.0 & 1 \\
     \\
    EFV && 0.01 & 5 && 0.1 & 2 \\
    ETR && 0.1 & 4 && 10.0 & 4 \\
    NVP && 10.0 & 1 && 0.1 & 4 \\
    RPV & \multicolumn{6}{c}{\textit{Removed from analysis due to low sample count}} \\
    \bottomrule
\end{tabular}
\end{center}

The CNN architecture used for DRM coverage comparison in section 9 was first published in \citep{steiner2020drug}. First, we tried to train CNN models using the R script provided by the authors on our datasets but, unfortunately, the script exited with errors and was not usable. Instead of debugging the provided script, which would already pose the risk of changing the training environment compared to the original publication, we decided to implement the same architecture using PyTorch. This course of action provided the additional benefit that the training of both neural network architectures, CNN and CMKN, used the same framework and followed the same standards which improves comparability of the methods. The first layer was an embedding layer that took the nominal encoded sequences and learned a three-dimensional embedding. Afterwards two one-dimensional convolutional layers with ReLu activation were used. Both convolutional layers had a kernel size of 9 with 32 filters. A one-dimensional MaxPooling layer with a pooling size of 5 was used between the two convolutional layers. The flatted output of the second convolutional layer was fed into fully-connected layers for the classification. CNN models were trained using PyTorch's implementation of the ADAM algorithm and class-balanced loss as introduced in \citep{cui2019class}. Training lasted for 200 epochs and no early-stopping was used. The learning rate started at $0.1$ and was dynamically adjusted using the \textit{ReduceLROnPlateau} method. We did not include the performance of the CNN architecture by Steiner et al. into our main manuscript due to the non-competitive performance on our datasets (i.e., they reached a mean MCC of 0.1). However, we included the description here since Steiner's models are used in section 9. The CNN results shown in the main manuscript were achieved with a network resulting from a simple ablation test, i.e., we replaced the kernel layer of an CMKN model with a standard convolutional layer with the same parameters.

\subsection{Splice Site Prediciton}
\subsubsection{CMKN}

In contrast to the HIV experiments, we only fixed two hyperparameters in the CMKN models used for splice site prediction: $\alpha$ was fixed to 1 and $\beta$ was fixed as described in Section \ref{sec:AppBeta}. The other hyperparameters were optimized with a grid search using the following values:
\begin{align}
k &\in \{ 2, 3, 4, 5 \} \nonumber \\
\sigma &\in \{ 1, 2, 4, 8, 16 \} \nonumber \\
\#\textrm{anchors} &\in \{ 15, 30, 45 \} \nonumber 
\end{align}
The optimal parameter choice for each splice site dataset can be found in Table \ref{tab:splice_bench}.

All CMKN models were trained using PyTorch's implementation of the ADAM algorithm and class-balanced loss as introduced in \citep{cui2019class}. Training lasted for 200 epochs and no early-stopping was used. The learning rate started at $0.1$ and was dynamically adjusted using the \textit{ReduceLROnPlateau} method.

\section{Comparing the Interpretation Capabilities of CMKNs}

\begin{figure}
    \centering
    \includegraphics[width=0.7\textwidth]{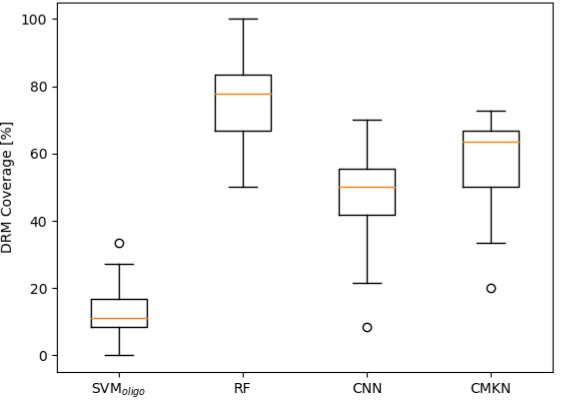}
    \caption{Number of identified drug resistance mutation positions (DRMs) by different models. \textit{Post-hoc} methods were needed to calculate feature importances for some models: For convolutional neural network (CNN) classifiers, we applied a permutation feature importance analysis and for the random forest (RF) classifiers, we calculated impurity-based feature importances. Position importances were directly assessable for CMKNs and oligo kernel SVMs ($\textrm{SVM}_{\textrm{oligo}}$). CMKNs performed similar to CNNs and RFs. All three significantly outperformed $\textrm{SVM}_{\textrm{oligo}}$. The center line of each box indicates the median. The height of the boxes represents the inter quartile range (IQR) with the upper and lower whiskers set to 1.5 times the IQR. Outliers are depicted by circles.}
    \label{fig:drms}
\end{figure}

\begin{figure}
    \centering
    \includegraphics[width=\textwidth]{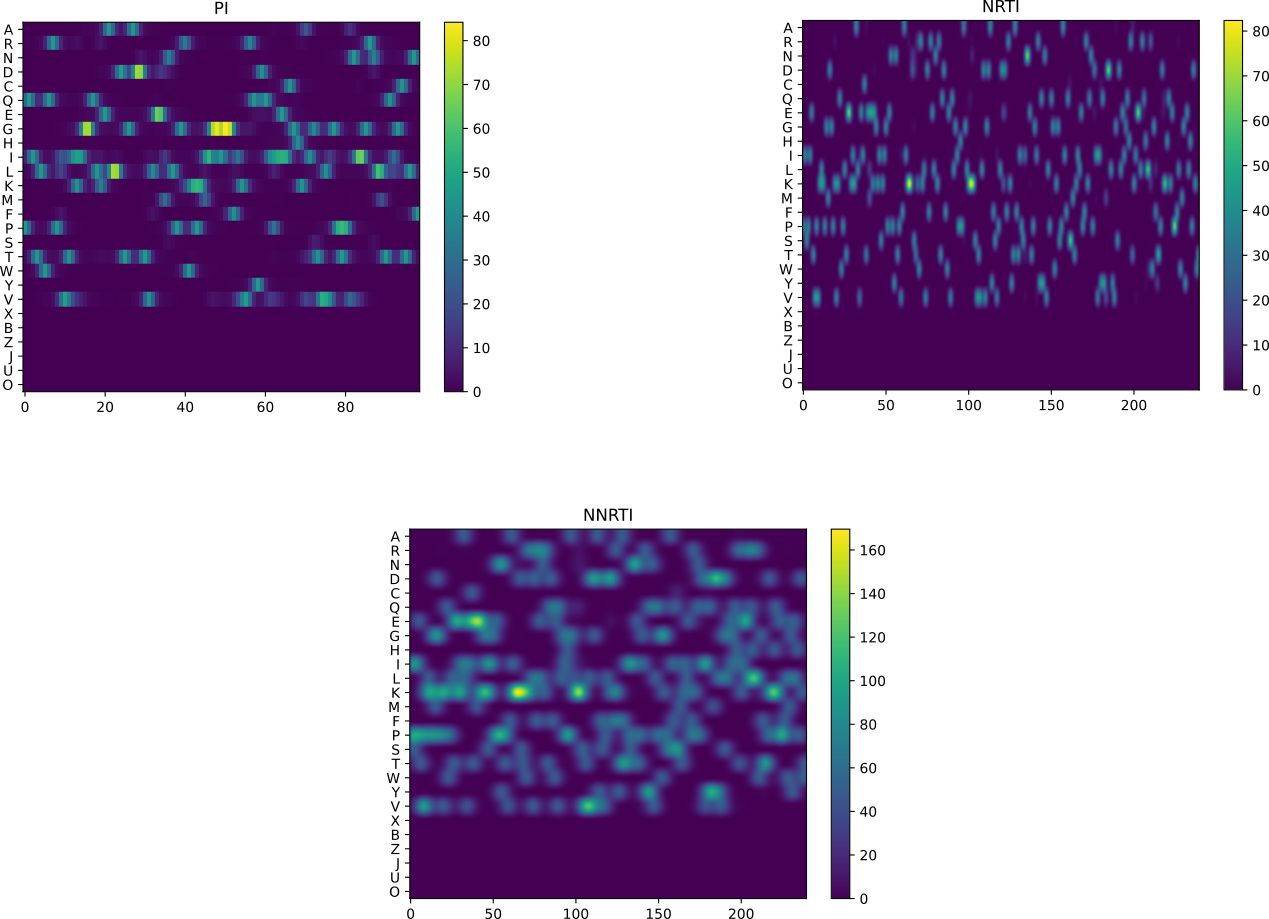}
    \caption{Interpretation of the support vector machines utilizing the oligo kernel ($\textrm{SVM}_{\textrm{oligo}}$). Details about the calculation of these matrices can be found in the original manuscript \citep{meinicke2004oligo}. $\textrm{SVM}_{\textrm{oligo}}$ had difficulties identifying drug resistance mutation positions (DRMs) and, due to it's limitation to $k$-mers, focused on single amino acids for each DRM.}
    \label{fig:oligo_m}
\end{figure}

Comparing the quality of interpretation capabilities of different methods is a difficult problem, due to the ethical and philosophical aspects of interpretation which are hard to express mathematically \citep{lipton2018mythos,morley2020ethics}. Nevertheless, we performed some quantitative comparisons of the global interpretation between the used inherently interpretable methods (i.e., CMKNs and oligo kernel SVMs) as well as published \textit{post-hoc} interpretations of non-interpretable methods (impurity-based feature importance for RFs \citep{raposo2020random} and permutation feature importance analysis for CNNs \citep{steiner2020drug}) on the HIV prediction task. For all methods, the 20 most important positions were identified and the percental coverage of drug resistance mutation positions (DRMs) using these 20 positions was calculated. The results can be found in Figure \ref{fig:drms}. We found that CMKNs performed similar to \textit{post-hoc} methods applied on CNNs and RFs when used to identify DRMs. All three methods significantly outperformed oligo kernel SVMs. However, the \textit{post-hoc} methods were limited to identifying positions without compositional information of the mutations. That lead to an incomplete picture of the biological process and therefore decreases the utility of the interpretation provided by \textit{post-hoc} methods. Only CMKNs and oligo kernel SVMs were able to provide the complete picture with positional and compositional information of the drug resistance mutations. We used the procedure described by Meinicke and colleagues in the original manuscript to visualize the learned oligo functions from the trained SVMs \citep{meinicke2004oligo}. The result is presented in Figure \ref{fig:oligo_m}. Due to the fact that the oligo kernel is limited to discrete $k$-mers, the compositional variability of DRMs is not identified by oligo kernel SVMs. For identified DRMS, the visualization focuses on single amino acids.

Additionally to global interpretations, CMKN models can also provide local interpretations. Comparing local interpretations cannot be achieved without a user study, since their utility is closely coupled with the intended user. Performing a user study with medical practitioners is a time consuming task, which is out of the scope of this manuscript. Nevertheless, evaluating local interpretations with the intended users is an important research question and will be investigated in future work.

\section{CMKN Model Interpretability for Splice Site Prediction}

\begin{figure}
    \centering
    \includegraphics[width=\textwidth]{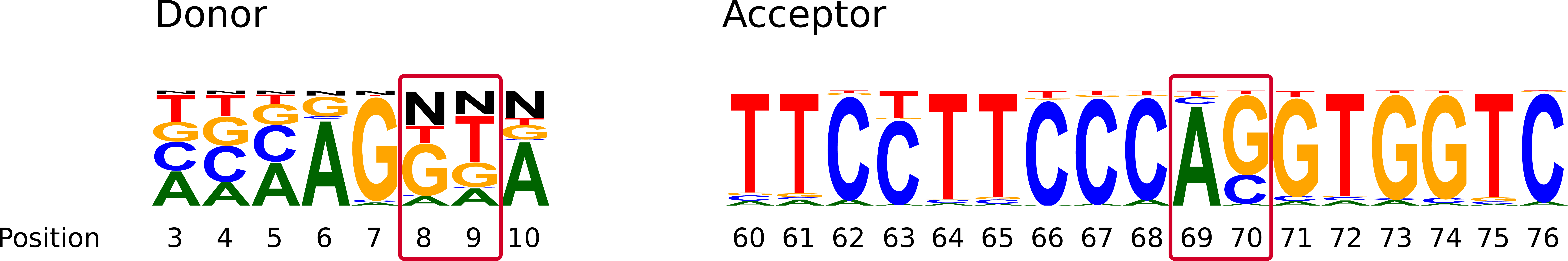}
    \caption{The learned sequence motifs of a CMKN model trained on the NN269 splice site prediction benchmark. The left motif was learned on the donor dataset. The right motif was learned on the acceptor dataset. The splice site is indicated by a red box.}
    \label{fig:interpretation_splice}
\end{figure}

We also analyzed how well CMKN models can retrieve the relevant motifs from the splice site prediction task. For this purpose, we retrained a model on the two NN269 training sets (donor and acceptor). We selected motif lengths that are able to capture the compositional variability that contributes to real splice sites in contrast to the decoy sequences. Therefore, the model trained on the donor dataset used a motif length of 8 and the model trained on the acceptor dataset used a motif length of 17. Since there is no positional variability in the splice site benchmarks, we only investigated the learned motifs. Figure \ref{fig:interpretation_splice} shows the motifs learned on the donor (left) and acceptor (right) dataset. Both splice site dimers (GT for donor sites and AG for acceptor sites) are correctly learned by CMKN models. Note that a high compositional variability in the learned motifs for these two positions is expected since they are exactly the same in real splice site sequences and decoy sequences. Therefore, the model does not put high distinction into these two positions resulting in less clear motifs. For the donor sequences, CMKN is able to identify the AG dimer that appears directly before the splice site and is highly predicative for a real splice site sequence. For the acceptor sequences, CMKN correctly learns the poly pyrimidine tract that precedes a real splice site together with the fact that a guanine directly after the splice site dimer is highly predicative for a real splice site. The learned motifs show that CMKN models are able to recover biologically meaningful sequence motifs from datasets.

\end{document}